\documentclass[11pt]{article}
\usepackage{vanilla_style}
\usepackage{orcidlink}
\usepackage{booktabs}

\usepackage[utf8]{inputenc}
\usepackage{hyperref}
\usepackage{xcolor}
\usepackage[font=small]{caption}

\usepackage{siunitx}
\usepackage{enumitem}

\definecolor{4-6}{RGB}{252,187,161}
\definecolor{7-9}{RGB}{252,146,114}
\definecolor{10-12}{RGB}{251,106,74}
\definecolor{13-15}{RGB}{222,45,38}
\definecolor{adults}{RGB}{165,15,21}

\definecolor{vgg19}{RGB}{188,189,220}
\definecolor{resnext}{RGB}{158,202,225}
\definecolor{bitm}{RGB}{107,174,214}
\definecolor{swsl}{RGB}{33,113,181}
\definecolor{swag}{RGB}{8,48,107}

\definecolor{midblue_errorK}{RGB}{55,126,184}

\usepackage[
style=apa
]{biblatex}

\author{Lukas S. Huber\textsuperscript{1,2, §} \orcidlink{0000-0002-7755-6926}\\
\and Robert Geirhos\textsuperscript{2} \orcidlink{0000-0001-7698-3187}\\
\and Felix A. Wichmann\textsuperscript{2} \orcidlink{0000-0002-2592-634X}\\
    }
  
\date{} 
\begin{document}

\title{The developmental trajectory of object recognition\\robustness: children are like small adults but unlike big deep neural networks\\\
 }

\maketitle

\begin{center}
\small{
\textsuperscript{1}\textit{Department of Psychology, University of Bern, Switzerland}\\
\textsuperscript{2}\textit{Neural Information Processing Group, University of Tübingen, Germany}\\
\textsuperscript{§}\textit{To whom correspondence should be addressed: \href{mailto:lukas.huber@psy.unibe.ch}{\texttt{lukas.s.huber@unibe.ch}}}
}

\end{center}

\begin{abstract}
\noindent In laboratory object recognition tasks based on undistorted photographs, both adult humans and Deep Neural Networks (DNNs) perform close to ceiling. Unlike adults', whose object recognition performance is robust against a wide range of image distortions, DNNs trained on standard ImageNet (1.3M images) perform poorly on distorted images. However, the last two years have seen impressive gains in DNN distortion robustness, predominantly achieved through ever-increasing large-scale datasets---orders of magnitude larger than ImageNet. While this simple brute-force approach is very effective in achieving human-level robustness in DNNs, it raises the question of whether human robustness, too, is simply due to extensive experience with (distorted) visual input during childhood and beyond. Here we investigate this question by comparing the core object recognition performance of 146 children (aged 4--15) against adults and against DNNs. We find, first, that already 4--6 year-olds showed remarkable robustness to image distortions and outperform DNNs trained on ImageNet. Second, we estimated the number of ``images'' children have been exposed to during their lifetime. Compared to various DNNs, children's high robustness requires relatively little data. Third, when recognizing objects children---like adults but unlike DNNs---rely heavily on shape but not on texture cues. Together our results suggest that the remarkable robustness to distortions emerges early in the developmental trajectory of human object recognition and is unlikely the result of a mere accumulation of experience with distorted visual input. Even though current DNNs match human performance regarding robustness they seem to rely on different and more data-hungry strategies to do so.

\vspace{5mm}

\end{abstract}

\section*{Introduction}

\emph{At a functional level, visual object recognition is at the center of understanding how we think about what we see} (\cite{Peissig_2007}, p. 76). Subjectively, visual object recognition typically appears effortless and intuitively easy to us; it is, however, an extremely difficult computational achievement: Arbitrary ``nuisance'' variables like object distance (size), pose and lightning potentially exert a massive influence on the proximal (retinal) stimulus, sometimes resulting in the very same distal stimulus (3-D object in a scene) to have a very different proximal stimuli. Conversely, for any given 2-D image on the retina---the proximal stimulus---there are an infinite number of potentially very different 3-D scenes---distal stimuli---whose projections would have resulted in the very same image (e.g., see \cite{pinto2008real,dicarlo2007untangling}).\footnote{At least if similarity is measured using standard image processing metrics like the Euclidean distance between, or correlation of, the images.} The human ability to recognise objects rapidly and effortlessly across a wide range of identity preserving transformations has been termed \emph{core object recognition} (see \cite{dicarlo2012does} for a review). The computational difficulty notwithstanding, human object recognition ability is not only subjective effortless but objectively often impressive (e.g., \cite{biederman1987recognition} or see \cite{logothetis1996visual,Peissig_2007,Gauthier_2015} for reviews). 

This computational complexity of (core) visual object recognition is also reflected in the long time it took computational models to reach human-level object classification accuracy---despite considerable, decade-long research efforts in vision science. It was not until 2012, when \textcite{krizhevsky2012imagenet} trained ``brain inspired'' deep neural networks (DNNs) on 1.3M natural images that computational models began to compete with humans in object recognition tasks. Today, DNNs are the state-of-the-art models in computer vision and surpass human performance on standard object recognition tasks such as image classification on the ImageNet dataset (e.g., see \cite{he2015delving}). It took even longer to obtain models that are not only performing well on natural, undistorted images similar to the training data but also, crucially, on more challenging datasets, so-called out-of-distribution (OOD) datasets containing for example image distortions that the models had never seen during training. This is precisely what humans excel on: robustly recognising objects even under hitherto unseen ``viewing conditions'' and distortions; in machine learning lingo humans show a high degree of OOD robustness. Even though some of these models have an innovative architecture and/or training procedure---such as CLIP \parencite{radford2021learning} or other variants of vision transformers \parencite{dosovitskiy2020image}---the most crucial feature to achieve human-like OOD robustness appears to be training on \textit{large-scale datasets} \parencite{geirhos2021partial}. While standard training on ImageNet includes 1.3M images, most models showing human-like OOD robustness are trained on much larger datasets---ranging from 14M (Big Transfer models; \cite{kolesnikov2020big}) to 940M images (semi-weakly supervised models; \cite{yalniz2019billion}) and even to a staggering 3.6B images \parencite{singh2022revisiting}. However, both architecture and data matter---vision transformers, e.g., trained on ImageNet are more robust than standard DNNs trained on ImageNet---but even standard DNNs trained on large-scale datasets (such as Big Transfer models) achieve remarkable robustness. This indicates that large-scale training may be \textit{sufficient} for OOD robustness in computational models. 

It remains an open question, however, whether large-scale experience is also \textit{neccessary} for robust core object recognition. This is precisely the question that we intend to answer with the present study: If large-scale exposure to visual input is indeed necessary to achieve a robust visual representation of objects, then we would expect human OOD robustness to be low in early childhood, and to increase with age due to continued exposure during lifetime. Alternatively, human OOD robustness might instead result from clever information processing and representation as well as suitable inductive bias \parencite{Mitchell_1980,Griffiths_2010}, achieving OOD robustness with comparatively little data. In this case, we would expect human OOD robustness to be already high in early childhood. Both hypotheses can be evaluated with developmental data.

Here we present a detailed investigation of the developmental trajectory of object recognition and its robustness in humans from age four to adolescence and beyond. We believe that resolving the competing hypotheses presented above may be relevant for understanding important aspects of both machine and human object recognition: in terms of machine vision, it is unclear whether large-scale training is the only way to achieve robustness---if children were able to achieve high robustness with little data, this would indicate that the limit of data-efficient robustness has not yet been reached. In terms of human vision, on the other hand, the developmental trajectory of object recognition robustness is still a puzzle with many missing pieces, limiting our understanding of the underlying processes and how they develop, as we will describe in the following section.

\subsection*{Development of object recognition}
Many cognitive abilities, like language or logical reasoning, mature with time; motor skills, too, take years to develop and be refined. What about our impressive object recognition abilities, particularly robustness to image degradations? Behavioral research investigating the \emph{development} of object recognition (robustness) in children (after 2 years of age) and adolescents is comparatively sparse, however. A number of reviews have pointed out the lack of such studies \parencite{nishimura2009development, smith2009fragments, rentschler2004development}. Clearly, the ventral visual cortex is subject to structural and functional changes from childhood, through adolescence and into adulthood (see \cite{grill2008developmental,RatanMurty-etal_2021} or \cite{klaver2011neurodevelopment} for a review). It has been shown that young children (5--12 years of age) already show adult-like category-selectivity for objects in the ventral visual cortex \parencite{scherf2007visual, golarai2010differential} and that the magnitude of retinotopic signals in V1, V2, V3, V3a and V4 are approximately the same in children as in adults \parencite{conner2004retinotopic}. In addition, contrast sensitivity in V1 and V3a also appears to reach adult-level by the age of seven \parencite{ben2007contrast}. These findings indicate that at least neural prerequisites for visual object recognition are in place at a comparatively early age.\footnote{Note that this is not the case for face-selective regions, which continue to develop well into adolescence \parencite{grill2008developmental}.} 

Most available behavioral data stems from children younger than two years. In those first two years of development, there are two major developmental changes. First, children start to use abstract representations of global shape rather than local features to recognise objects \parencite{augustine2011parts, pereira2009developmental, smith2003learning}. This change enables adult-like performance in simple object recognition tasks and is thought to facilitate generalization and increase the robustness of object recognition \parencite{son2008simplicity}. Second, children start to use object shape as the crucial property to generalize names to never before seen objects---a tendency termed shape bias (e.g., see \cite{landau1988importance}). An empirical study suggests that these two changes are developmentally connected, so that the ability to form abstract representations of global object shape precedes the shape bias \parencite{yee2012changes}. To our knowledge, only one study has systematically investigated the development of object recognition after the age of two. \textcite{bova2007development} have shown a progressive improvement of visual object recognition abilities in children from 6 to 11 years of age as measured by a battery of neuropsychological tests.\footnote{Tests used included the Efron Test, Warrington’s Figure-Ground Test, the Street Completion Test, the Poppelreuter-Ghent Test, a selection of stimuli from the Birmingham Object Recognition Battery, and a series of color photographs of objects presented from unusual perspectives or illuminated in unusual ways.} They report that simple visual abilities (such as shape discrimination) were already mature at the age of six whereas more complex abilities (such as the recognition of objects presented in a hard-to-decode way) tended to improve with age.\footnote{Note that most tests were very different from the psychophysical task we employed in this study. The tests most similar to the task used here are those administered to asses the ability to recognise the structural identity of an object even when its projection on the retina is altered---perceptual categorisation as meassured by the Street Completion Test, the Poppelreuter-Ghent Test, and the identification of color photographs of objects viewed from unusual perspectives and presented under unusual lighting conditions. However, all of these test only consist of a small number of stimuli (11, 13 or 44 respectively), did not apply parameterised distortions to real photographs and did not implement a limited stimulus-presentation duration---and thus are not suitable as a rigorous psychophysical assessment of core object recognition.} However, this study did not use stimuli typically used to assess object recognition in adults or DNNs, preventing any quantitative comparisons to be made---something we attempt to remedy with the present study (but see footnote~\footref{foot:ayzenberg}). In the present study, we investigate how well children of different age groups (4--6, 7--9, 10--12, 13--15) are able to recognise objects in 2D images at different levels of difficulty (degree of distortions) in order to trace the developmental trajectory of human object recognition robustness.

\section*{Method}

\subsection*{General}
\label{sec:general}

The methods used in this study are adapted from a series of psychophysical experiments conducted by Geirhos and colleagues (\citeyear{geirhos2018generalisation, geirhos2019imagenet}). The paradigm is an image-category identification task to compare human observers and DNNs as fairly as possible. Images are presented on a computer screen and for each image observers are asked to choose the corresponding category as quickly and accurately as possible. Concerning the fairness of the comparison between humans and DNNs one aspect needs to be highlighted: Standard DNNs are typically trained on the ImageNet (ILSVRC) database \parencite{russakovsky2015imagenet}, which contains about 1.3 million images grouped into 1,000 fine-grained categories (e.g., over a hundred different dog breeds). However, human observers categorise objects most quickly and naturally at the entry-level, which is very often the basic-level e.g., \texttt{dog} rather than \texttt{German shepherd} \parencite{Rosch_1973,Rosch_1979}. In order to account for this discrepancy and to provide a fair comparison, Geirhos et al.\ used a mapping from 16 human-friendly entry-level categories to their corresponding ImageNet categories based on the WordNet hierarchy \parencite{miller1995wordnet}.

We adapted the following aspects of the original Geirhos et al.\ studies to make the paradigm more suitable to test young children: We introduced a certain degree of gamification, added more breaks, and did not force the children to respond within 1500 msec after stimulus offset to avoid undue stress. After each block of 20 trials children were free to either quit the experiment or continue with another block. Compared to Geirhos et al., we slightly increased the stimulus presentation duration from 200 ms to 300 ms and only used stimuli which were correctly recognised by at least two adults in the previous studies. We employed a between-subject design to test participants on two different types of distortions: binary salt-and-pepper noise and so-called ``eidolon'' distortions \parencite{koenderink2017eidolons}. In an additional experiment we also used texture-shape cue-conflict stimuli as in \textcite{geirhos2019imagenet}. In what follows we first provide a description of the \hyperref[sec:procedure]{procedure}, the introduced \hyperref[sec:gamification]{gamification} and the employed \hyperref[sec:stimuli]{stimuli}. We then proceed by giving details on the tested \hyperref[sec:participants]{participants} (children, adolescents, adults), the \hyperref[sec:apparatus]{experimental setup}, and the evaluated \hyperref[sec:models]{DNNs}.

\subsection*{Procedure}
\label{sec:procedure}
\begin{figure}[!ht]
   \begin{center}
        \includegraphics[scale=.8]{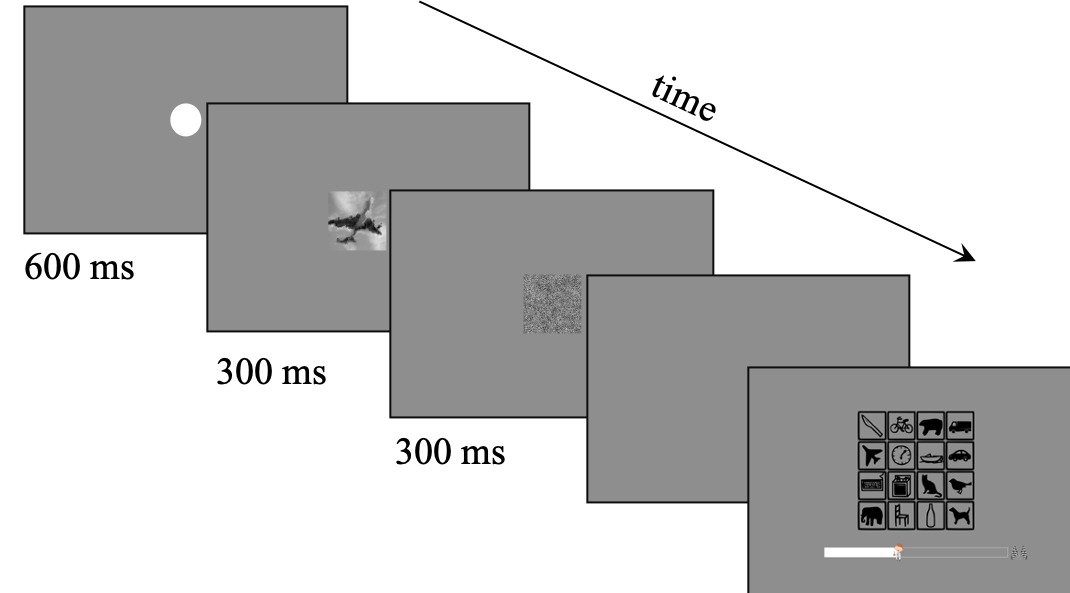}
    \end{center}
       \caption{ Schematic of a trial. After the attention grabber clock had expired (600 ms), the target image was presented for 300 ms, followed immediately by a full-contrast pink noise mask (1/$f$ spectral shape) of the same size and duration. Following the mask, participants had unlimited time to make their choice employing the physical response surface. Note, however, that participants were instructed to respond as quickly and accurately as possible. After the participant responded, the response surface was shown on screen and the experimenter clicked on the icon corresponding with the participant's response. Icons on the response screen represented the 16 entry-level categories---row-wise from top to bottom: \texttt{knife, bicycle, bear, truck, airplane, clock, boat, car, keyboard, oven, cat, bird, elephant, chair, bottle, dog}. Below the response surface there is a gamified progress-bar indicating the degree to which the current block has been completed.}
       \label{fig:trial}
\end{figure}

Each trial consisted of several phases. First, we presented an attention grabber inspired by an expiring clock (solid white circle which empties itself within 600 ms) in the center of the screen. We chose a moving stimulus instead of a more commonly used fixation cross in order to compensate for possible weaker attention in children. Second, the target image was shown in the center of the screen for 300 ms, followed by a full-contrast pink noise mask (1/$f$ spectral shape) of the same size and duration to prevent after-images and limit internal processing time. Next, the screen turned blank and participants were required to indicate their answers. They did this by physically pointing to one of 16 icons corresponding to the 16 entry-level categories on a laminated DIN A4 sheet arranged in a $4\times4$ grid (icon size: $3\times3$ cm). We chose this physical response surface mainly for time efficacy (having 4-year-olds handle a computer mouse by themselves can be a lengthy and somewhat unreliable undertaking). Next, the 16 icons appeared on screen, and the experimenter recorded the response provided by the child using a wireless computer mouse. As in the experiments conducted by Geirhos et al., our icons were a modified version of the ones from the MS COCO website (\url{https://cocodataset.org/#explore}). Figure \ref{fig:trial} shows the schematic of a trial.

All participants were tested in a separate, quiet room---either in their school (children, adolescents) or at home (adults). The experimental session started with the presentation of example images. For each category we showed a prototypical example image in the center of the screen and asked participants to name the depicted object. The correct category was indicated by the subsequent presentation of the corresponding category icon. After completing all 16 examples, participants completed ten practice trials on undistorted color images (no overlap with stimuli from experimental trials). Extremely rarely some of the youngest children failed on two or more images and had to complete another round of ten practice trials. Before the experimental trials started, a single distorted image (matched for the given experimental distortion) was shown and a short story-like explanation was given to justify why some of the subsequent images would be distorted.\footnote{Eidolon: Someone left the images in the rain; that is why some of them are blurry. Noise: Somebody spilled salt and pepper; that is why some of them look a bit strange. Cue-Conflict: Someone left the images in the beating sun; that is why some of them stuck together.} Experimental trials were arranged in blocks containing 20 trials each. After each block, participants received feedback and were asked whether they would like to continue, have a break or terminate the session. Adults were not explicitly asked if they want to terminate the session---but of course \textit{all} participants were informed at the outset that they could abort the experiment at any given time. Participants could complete a maximum of 16 blocks (320 images) in the eidolon and salt-and-pepper experiments and 20 blocks (400 images) in the cue-conflict experiment.

\subsection*{Gamification}
\label{sec:gamification}
In order to increase motivation and make the experiment more appealing to children, we gamified several aspects of the experiment. In the beginning, participants could choose one of four characters (matched for gender) corresponding to four different roles: spy, detective, scientist or safari guide. The chosen character had to undergo a training session designed to improve her or his crucial skill. What the participants did not know was that the crucial skill---identifying objects as quickly and accurately as possible---was the same for all characters. After each trial, the chosen character was displayed at the foremost position of a progress bar indicating how far the participant had progressed in the current block (level). After each block, participants were provided with feedback designed to be perceptually similar to the display of a game score in an arcade game. There were three different types of scores. Participants received 10 \texttt{coins} as a reward for a finished block (not performance related). Additionally, for every two correctly recognised images, they received a \texttt{star} (performance-related). If they scored more than eight stars, they earned a special \texttt{emblem} matched for the chosen story character.\footnote{The emblems were sunglasses for the spy, a magnifying glass for the detective, a microscope for the scientist, and a camera for the safari guide.} Figure~\ref{fig:gamification} in the Appendix~\hyperref[app:gamification]{A.1} visualises different gamified elements.

\subsection*{Stimuli}
\label{sec:stimuli}

\subsubsection*{Salt-and-pepper noise and eidolon distortion}

\begin{figure}[!ht]
     \begin{center}
     \begin{subfigure}[b]{0.22\textwidth}
         \centering
         \includegraphics[width=\textwidth]{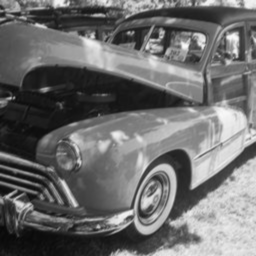}
         \caption{Noise level = 0}
         \label{fig:3a}
     \end{subfigure}
     \hfill
     \begin{subfigure}[b]{0.22\textwidth}
         \centering
         \includegraphics[width=\textwidth]{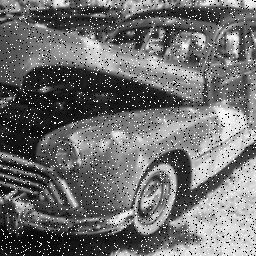}
         \caption{Noise level = 0.1}
         \label{fig:3b}
     \end{subfigure}
     \hfill
     \begin{subfigure}[b]{0.22\textwidth}
         \centering
         \includegraphics[width=\textwidth]{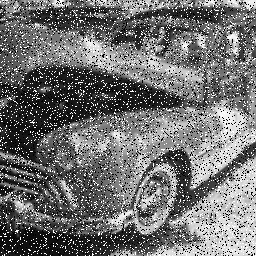}
         \caption{Noise level = 0.2}
         \label{fig:3c}
     \end{subfigure}
     \hfill
     \begin{subfigure}[b]{0.22\textwidth}
         \centering
         \includegraphics[width=\textwidth]{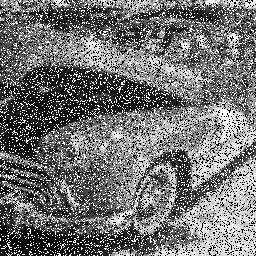}
         \caption{Noise level = 0.35}
         \label{fig:3d}
     \end{subfigure}
     \hfill
     \begin{subfigure}[b]{0.22\textwidth}
         \centering
         \includegraphics[width=\textwidth]{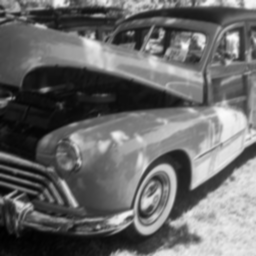}
         \caption{Reach level = 0}
         \label{fig:3e}
     \end{subfigure}
     \hfill
     \begin{subfigure}[b]{0.22\textwidth}
         \centering
         \includegraphics[width=\textwidth]{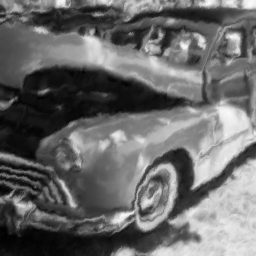}
         \caption{Reach level = 4}
         \label{fig:3f}
     \end{subfigure}
     \hfill
     \begin{subfigure}[b]{0.22\textwidth}
         \centering
         \includegraphics[width=\textwidth]{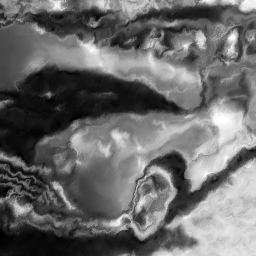}
         \caption{Reach level = 8}
         \label{fig:3g}
     \end{subfigure}
     \hfill
     \begin{subfigure}[b]{0.22\textwidth}
         \centering
         \includegraphics[width=\textwidth]{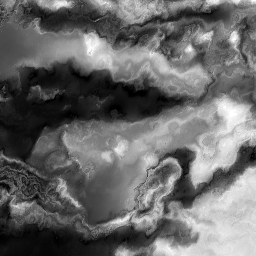}
         \caption{Reach level = 16}
         \label{fig:3h}
     \end{subfigure}
     \hfill
    \end{center}
    \caption{Systematic degradation of images in the salt-and-pepper noise (\textbf{a-d}) and the eidolon (\textbf{e-h}) experiments. Note that even though different degradation levels are shown for the same image, participants never encountered the same initial image multiple times.}
    \label{fig:stimuli}
\end{figure}

As mentioned above we used images from 16-class-ImageNet \parencite{geirhos2018generalisation}. A subset of 521 stimuli---stimuli that were correctly classified by at least two adults in prior experiments---served as a starting point for the present study. We chose this subset because we feared that children's motivation might be weaker compared to that of adults and wanted to avoid frustrating children with stimuli that even adults are unable to recognize. We then randomly sampled 320 images (20 for each of the 16 categories) to be manipulated in the next step. For both experiments (eidolon and noise), we manipulated the images to four different degrees, resulting in 4 different difficulty levels per experiment. For the eidolon experiment, we used the eidolon toolbox \parencite{koenderink2017eidolons} with the following settings: $grain = 10$, $coherence = 1$ and four different $reach$ $levels$ corresponding to the four difficulty levels ($0, 4, 8, 16$). The higher the reach level, the more distorted the images are and the more difficult it is to recognise them. In the noise experiment, a certain proportion of pixels were either set to a gray value of 1 (white) or 0 (black). This manipulation is often referred to as \emph{salt and pepper noise}. The four difficulty levels in this experiment corresponded to 4 different proportions of flipped pixels (0, 0.1, 0.2 or 0.35). For example, 0.2 means that 20\% of the pixels are switched while 80\% of the pixels remain untouched. For simplicity, we use the term ``difficulty level'' to refer to both: the different reach levels of the eidolon experiment and the different noise levels of the salt-and-pepper noise experiment. It is important to note, however, that the difficulty levels were not precisely matched between conditions as can be seen in the results. Figure \ref{fig:stimuli} displays an example image to which both distortions were applied.

For each difficulty level we randomly selected five images per category to be distorted. Note that in both experiments the lowest difficulty level (either reach level or proportion of switched pixels equals 0) can be interpreted as only a gray scale transformation of the original color images.\footnote{This is true for the noise images, however, this was not fully true for the eidolon transformations: As can be seen in sub-figure \hyperref[fig:3a]{3a}, the sharpness decreased a little bit compared to the original images---an unforseen result of the eidolon toolbox.} Next, we divided the 320 images into four chunks of 80 images each. Such a chunk features five images per category and 20 images per difficulty level. From each chunk, we created four blocks of 20 images each, resulting in 16 blocks that we later used in the experiment. To minimize predictability, individual blocks of 20 images were not balanced for categories (i.e. a block could contain a variable number of images from a given category). However, each block was balanced for difficulty levels (five images per difficulty level). To keep the participant's motivation as high as possible, we pseudo-randomized the order of image presentation within each block during the experiment given the following constraint: The first and last three images of a block had to be easy to recognise (difficulty level one or two).  

\subsubsection*{Cue-conflict}
\label{sec:cueconflict}

\begin{figure}[!ht]
     \begin{center}
     \begin{subfigure}[b]{0.25\textwidth}
         \frame{\includegraphics[width=\textwidth]{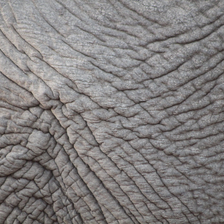}}
         \caption{Texture image}
         \label{fig:y equals x}
     \end{subfigure}
     \hfill
     \hspace{1.5cm}
     \begin{subfigure}[b]{0.25\textwidth}
         \centering
         \frame{\includegraphics[width=\textwidth]{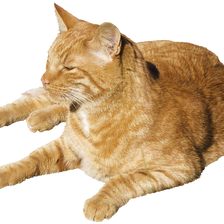}}
         \caption{Content image}
         \label{fig:three sin x}
     \end{subfigure}
     \hfill
     \hspace{1.5cm}
     \begin{subfigure}[b]{0.25\textwidth}
         \centering
         \frame{\includegraphics[width=\textwidth]{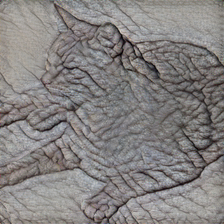}}
         \caption{Texture-shape cue conflict}
         \label{fig:five over x}
     \end{subfigure}
     \hfill
        \end{center}
        \caption{Stimuli generation for cue-conflict images. Initial images (\textbf{a}, \textbf{b}) used for style transfer \parencite{gatys2016image} in order to create texture-shape cue conflict stimuli (\textbf{c}) employed in the cue-conflict experiment. Note that participants never encountered texture or content images. They only encountered texture-shape cue conflict images and original images that were similar to content images but featured natural backgrounds. Figure adapted from \textcite{geirhos2019imagenet}.}
        \label{fig:cue-conflict}
\end{figure}

In the cue-conflict experiment we used a subset of images as used in \textcite{geirhos2019imagenet}. These 224$\times$224 pixel cue-conflict images are designed to have a conflict between two cues, namely, object \textit{shape} and object \textit{texture}, for example, the shape of a cat combined with the texture of elephant skin (see Figure \ref{fig:cue-conflict}). The stimuli were created using the style transfer method \parencite{gatys2016image}, whereby the content of an image (shape) is combined with the appearance of another image (texture) using a DNN-based approach. From the 1280 cue-conflict images created by \textcite{geirhos2019imagenet} we sampled 240 images (15 per category) to use in this experiment. We included 160 original colour images (10 per category) as a baseline and to help keep the task intuitive for the children (sampled from the 521-image subset of 16-class-ImageNet as described above). The whole sample of 400 images was split into five chunks of 80 images each---32 original images (2 per category) and 48 cue-conflict images (3 per category). As in the other two experiments, we created four blocks (20 stimuli each) from each chunk, resulting in 20 blocks that we later used in the experiment. The selection of the images was again not balanced regarding categories but for difficulty levels (i.e., each block contained eight original and 12 cue-conflict images). Again, the order of image presentation in the experiment was pseudo-random with the following constraint: The first and last three images had to be original but not cue-conflict images.

\subsection*{Participants}
\label{sec:participants}
In total we collected 23,474 trials from a sample of 146 children and adolescents (4--15 years) and nine adults. Participants were assigned to one of three experiments: \textit{Noise} (48 children and three adults, 60\% female), \textit{Eidolon} (46 children and three adults, 45\% female) and \textit{Cue-Conflict} (52 children and three adults, 45\% female). Further descriptive information about the sample and observations is presented in Table~\ref{tab:demo} in the Appendix~\hyperref[app:demo]{A.2}. We recruited children from 17 different schools in Bern (Switzerland). The adult sample was recruited through private contacts. All participants reported normal or corrected to normal vision, provided (parental) written consent, and were tested in accordance with national and international norms that govern research with human participants. The study was approved by the institutional ethical review board of the University of Bern (no.~2020-08-00003). As a token of appreciation for their participation, children received a book of their choice. Only one child decided to cancel the study right after completing practice trials.

\subsection*{Apparatus}
\label{sec:apparatus}
Programming and stimulus presentation were realized with Python (version 3.8.2) on a Lenovo Thinkpad T490s (Quad core CPU i5-8365U, Intel UHD 620 graphic card) running Linux Mint 20 Ulyana. We programmed the experiment's interface with the Psychopy library (\cite{Peirce2019}; version 2020.2.4). The 14” screen (356 mm diagonal) had a spatial resolution of $1920\times1200$ pixels at a refresh rate of 120 Hz. Measured luminance of the display was 361.4 cd/m\textsuperscript{2} and gamma set to 2.2. Images were presented at the center of the screen with a size of $256\times256$ pixels, corresponding, at a viewing distance of approximately 60 cm, to $4\times4$ degrees of visual angle.\footnote{This is only true for the eidolon and noise experiment. Due to an unnoticed cropping error image size in the cue-conflict experiment was $224\times224$ pixels, corresponding, at a viewing distance of approximately 60 cm, to only $3.5\times3.5$ degrees of visual angle. We do not think that this small change in absolute size had any influence on the data or results we report.} Note that viewing distance varied somewhat between participants due to children's agitation. For the whole experiment the background color was set to a gray value in the [0,1] range corresponding to the mean grayscale value of all images in the dataset of the particular experiment (eidolon: 0.452, noise: 0.459, and cue-conflict: 0.478).\footnote{To evaluate the mean grayscale value, images in the cue-conflict data set were converted to grayscale using \texttt{skimage.color.rgb2gray}.} All responses were recorded with a standard wireless computer mouse.

\subsection*{Models}
\label{sec:models}
In order to investigate the effect of dataset size on model robustness, we selected four representative models from the \href{https://github.com/bethgelab/model-vs-human/}{\texttt{modelvshuman}} Python-toolbox \parencite{geirhos2021partial}. The models were chosen according to the following criteria: (1.) in terms of training dataset size, they are separated by an approximate log unit each; (2.) to a certain degree, they are all derivatives of ResNet building blocks \parencite{he2015delving}; and (3.) within the class of models that satisfies the first two constraints, each of them is the \emph{very best performing model} in terms of OOD accuracy as evaluated on the \texttt{modelvshuman} benchmark---thus they are, as of now, some of the most robust DNNs and therefore the strongest DNN competitors for our human to DNN robustness comparison. According to these criteria, the following four models were chosen:

\begin{itemize}[leftmargin=2.3cm]
\item[$>1$M:] \textbf{ResNeXt}: a ResNeXt-101\_32x8d model by \textcite{xie2017aggregated} trained on 1.3M images;
\item[$>10$M:] \textbf{BiT-M}: a BiT-M model by \textcite{kolesnikov2020big} based on a ResNetV2-152x2 trained on 14M images;
\item[$>100$M:] \textbf{SWSL}: a SWSL model by \textcite{yalniz2019billion} based on a ResNeXt-101\_32x16d trained on 940M images; 
\item[$>1,000$M:] \textbf{SWAG}: a SWAG model by \textcite{singh2022revisiting} based on a RegNetY-128GF trained on 3.6B images.
\end{itemize}

Additionally, we decided to include one widely known DNN, \textbf{VGG-19} by \textcite{simonyan2014very}, for comparison purposes since it is based on a very simple architecture and it has been studied extensively in the past. 

We used a single feed-forward pass with $224\times224$ px RGB images except for the SWAG model which requires $384\times384$ px input. In this case, the images were scaled up to $384\times384$ px using PIL.Image.BICUBIC interpolation. For grayscale images (noise and eidolon experiment) all three channels were set to be equal to the grayscale image's single channel.

\section*{Results}
Recent machine learning models have seen tremendous gains in object recognition robustness, predominantly achieved through ever-increasing large-scale datasets. Here we ask whether human robustness, too, may simply be the result of extensive visual experience acquired during lifetime. If so, we would expect human robustness to be low in young children, and to increase over the years. To this end, we perform three comparisons between children of different age groups vs.\ adults and vs.\ DNNs. We first investigate the developmental trajectory of object recognition robustness (section~\hyperref[sec:robustness_development]{\mbox{developmental trajectory}}). Having found essentially adult-level robustness already in young children, we then perform a back-of-the-envelope calculation to estimate bounds on the amount of ``images'' that children can possibly have been exposed to (section~\hyperref[sec:envelope_calculation]{back-of-the-envelope calculation}). Finally, we investigate whether object recognition \emph{strategies} change over the course of development (section~\hyperref[sec:strategy_development]{strategy development}).

\subsection*{Developmental trajectory: human object recognition robustness develops early}
\label{sec:robustness_development}

\begin{figure}
    {\footnotesize
    \textbf{Accuracy} \hspace{7.8cm} \textbf{Normalized accuracy}}
    \begin{center}
    \captionsetup[subfigure]{labelformat=empty}
     \begin{subfigure}[b]{0.44\textwidth}
         \centering
         \includegraphics[width=\textwidth]{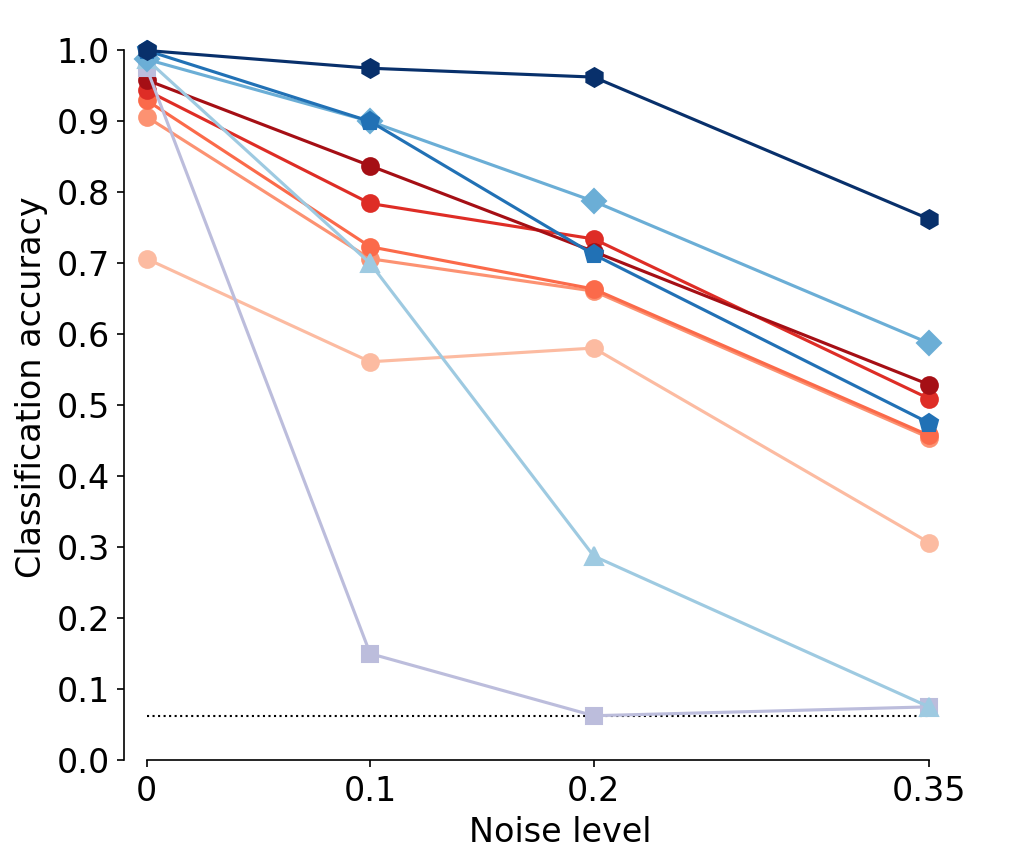}
         \caption{(b) Salt-and-pepper noise}
         \label{fig:five over x}
     \end{subfigure}
     \hfill
     \hspace{1.2cm}
     \begin{subfigure}[b]{0.44\textwidth}
         \centering
         \includegraphics[width=\textwidth]{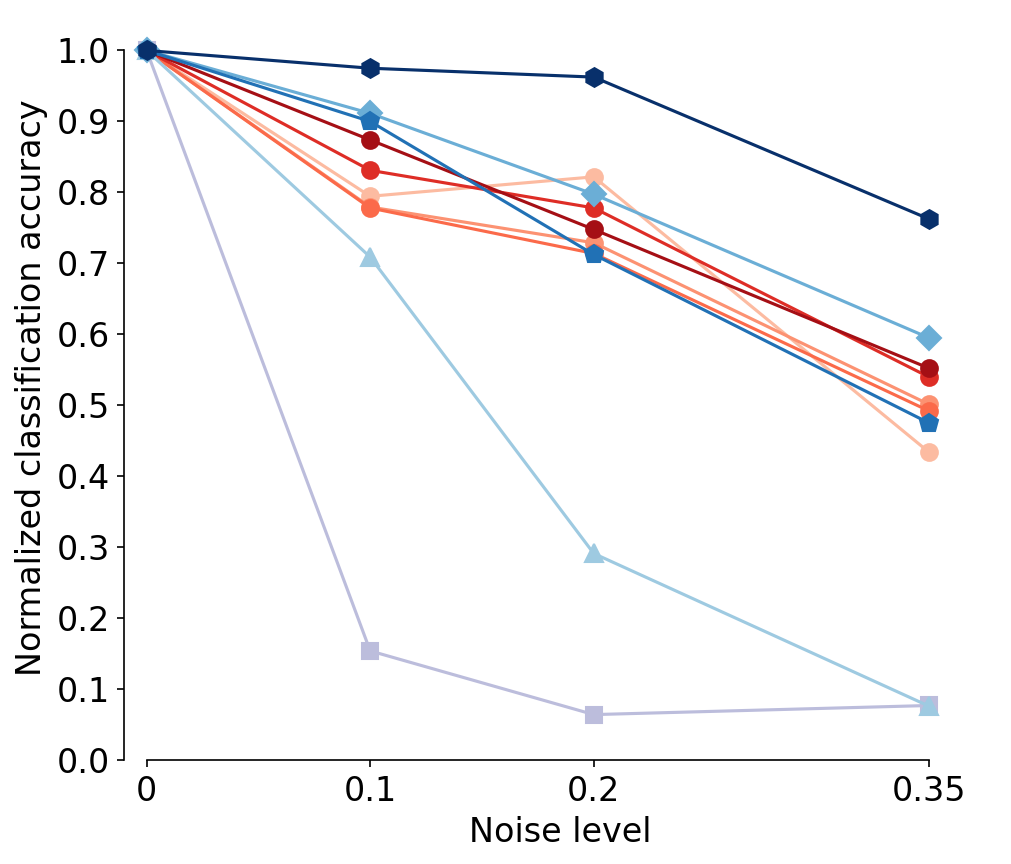}
         \caption{}
         \label{fig:five over x}
     \end{subfigure}
     \begin{subfigure}[b]{0.44\textwidth}
         \includegraphics[width=\textwidth]{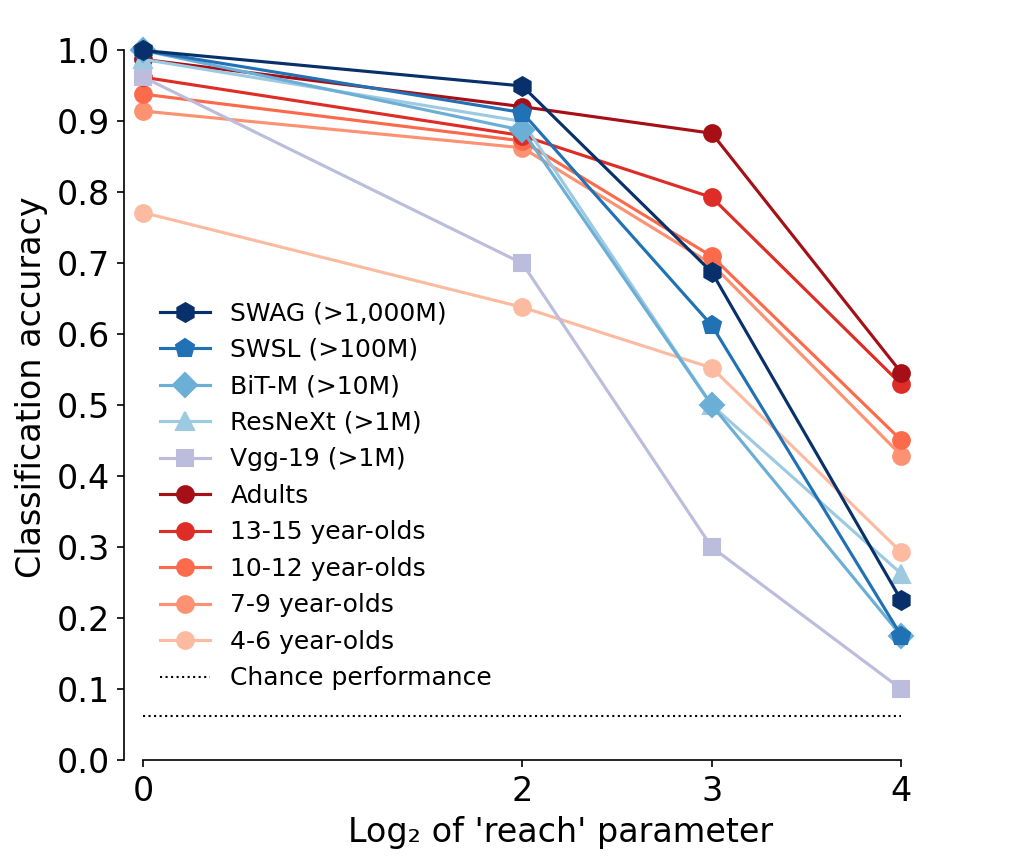}
         \caption{(a) Eidolon}
         \label{fig:five over x}
     \end{subfigure}
     \hfill
     \begin{subfigure}[b]{0.44\textwidth}
         \centering
         \includegraphics[width=\textwidth]{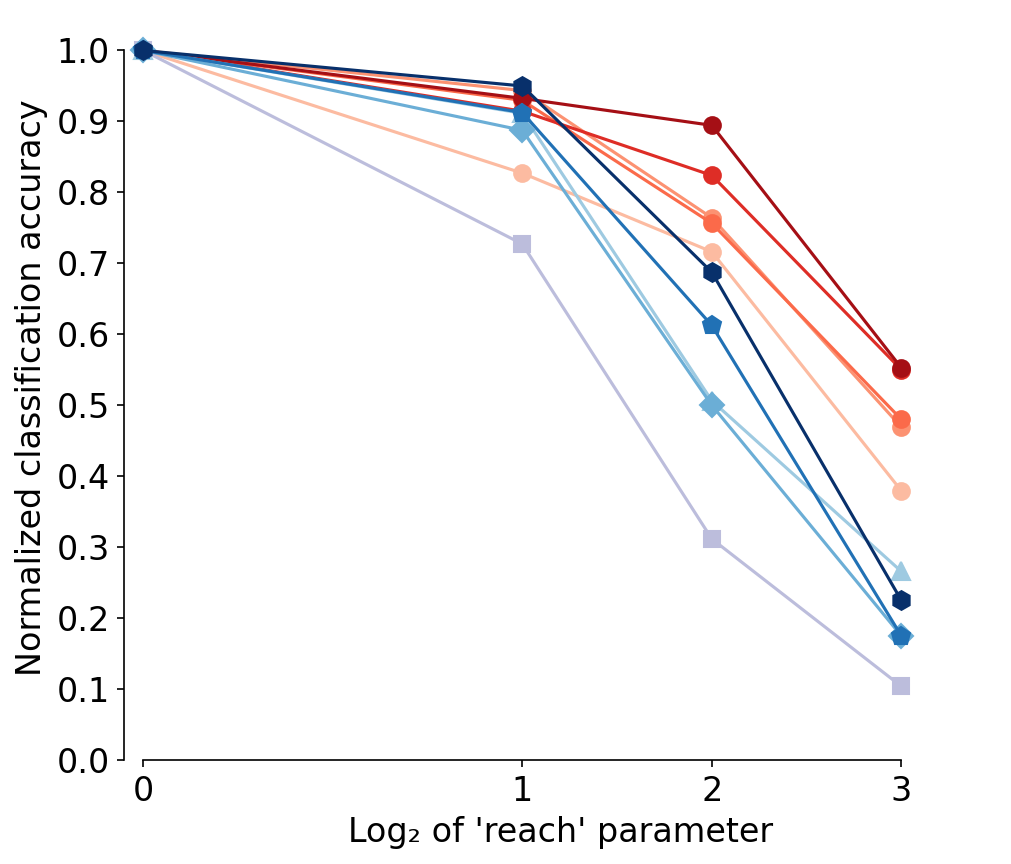}
         \caption{}
         \label{fig:five over x}
     \end{subfigure}
     \hfill
     \hfill
    \end{center}
    \caption{Classification accuracy (top 1) and normalized classification accuracy for different age groups and models. (\textbf{a}) shows the results for the salt-and-pepper noise and (\textbf{b}) for the eidolon experiment. The dotted line represents chance level performance of 6.25 \% (100 \% divided by the number of categories, which was 16). Normalized accuracy shows the change in accuracy relative to the initial accuracy at difficulty level zero of each age group or DNN, respectively.}
    \label{fig:accuracy}
\end{figure}

In order to assess the developmental trajectory of object recognition robustness, we measure classification accuracy depending on the amount of image degradation for two different experiments: salt-and-pepper noise and eidolons (visualised in Figure~\ref{fig:stimuli}). The results are shown in the left column of Figure~\ref{fig:accuracy}. In addition to classification accuracy, we plot normalized accuracy with respect to the initial accuracy at difficulty level zero since this makes it easier to disentangle the effects of initial accuracy and of a change in robustness (right column in Figure~\ref{fig:accuracy}).

First, looking at classification accuracy it can be seen that while the adults' performance is close to ceiling at difficulty level zero, there is a moderate decrease in accuracy as the difficulty level increases (circles, \textcolor{adults}{dark red}). However, even at difficulty level three, adults still demonstrate a fairly high accuracy, which is far above chance level. The robustness trajectories for DNNs differ dramatically (shades of blue and violet): older models trained on ImageNet ($>$1M images) are typically far below the human level (VGG-19; ResNeXt), while modern models trained on large-scale datasets ($>$10, $>$100 or even $>$1,000M images) are sometimes even above the human level, a finding consistent with \textcite{geirhos2021partial} who reported that the model-to-human gap in out-of-distribution distortion robustness has essentially closed. However, the developmental trajectory of robustness during childhood and adolescence has not been studied so far: Across both experiments (salt-and-pepper noise as well as eidolons), overall performance increases as a function of age. The biggest gain in performance is achieved between the groups of 4--6 year-olds (circles, \textcolor{4-6}{ light orange}) and 7--9 year-olds (circles, \textcolor{7-9}{darker orange}). That being said, across all age groups, there appears to be only a linear offset when compared to adults (who have a similar slope): relative to their performance level at zero noise or distortion, even 4--6 year-olds appear to have acquired essentially adult-like robustness---i.e., their relative (normalized) performance under noise is similar to that of adults (top right panel) or very nearly so (bottom right panel).

\begin{figure}[!ht]
     \begin{center}
     \begin{subfigure}[b]{0.35\textwidth}
         \centering
         \includegraphics[width=\textwidth]{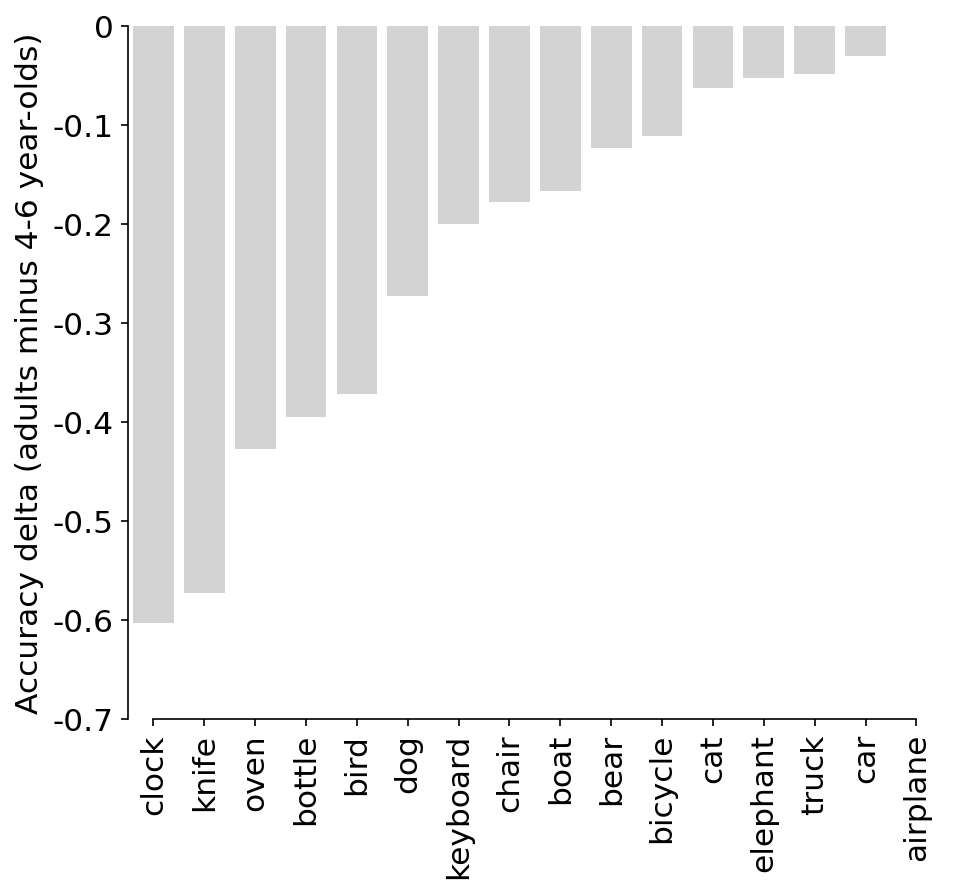}
         \caption{Class-conditional accuracy difference (adults minus 4-6-year-olds)}
         \label{subfig:delta}
     \end{subfigure}
     \begin{subfigure}[b]{0.55\textwidth}
         \centering
         \includegraphics[width=\textwidth]{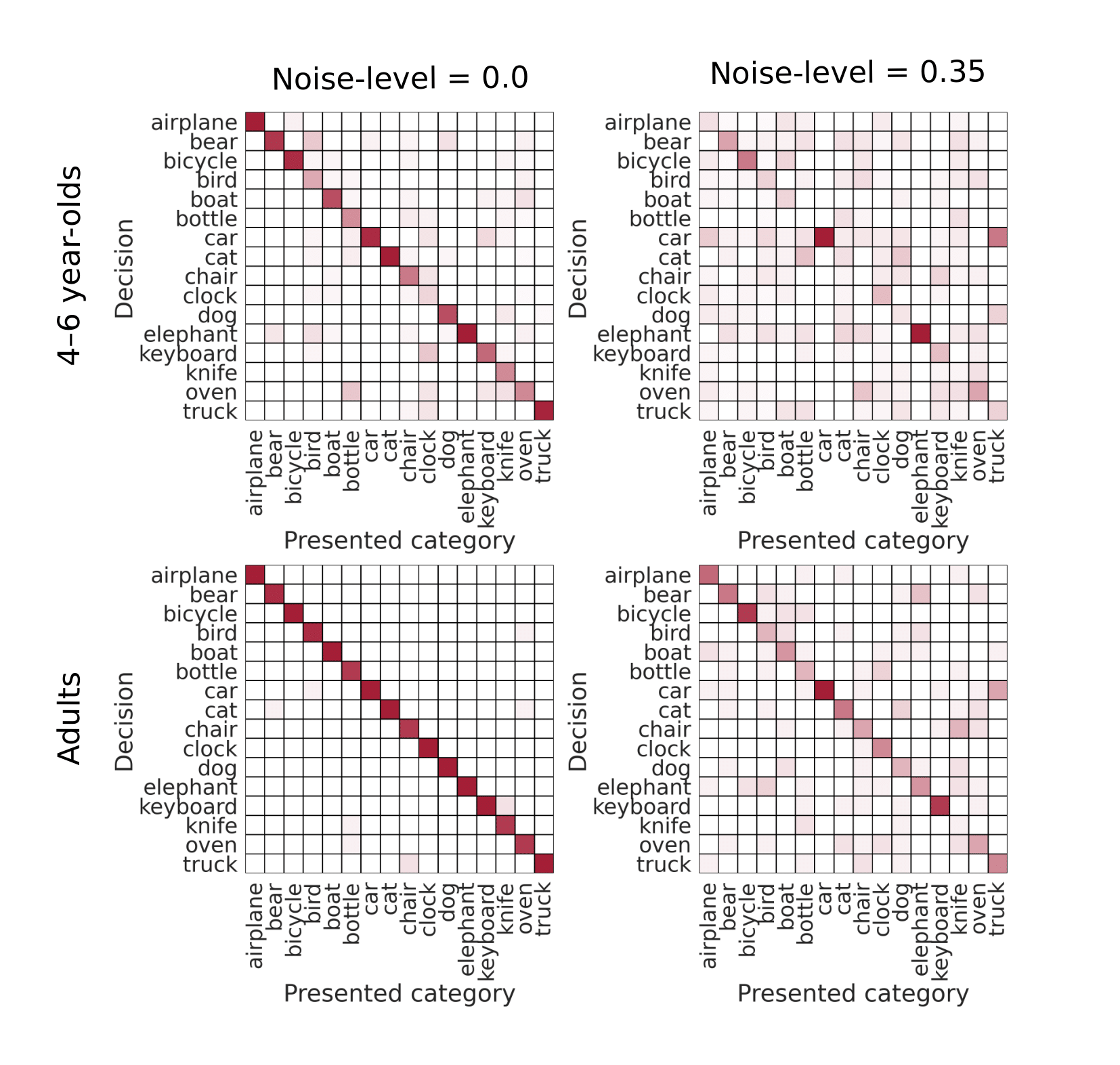}
         \caption{Confusion matrices}
         \label{subfig:confusion}
     \end{subfigure}
    \end{center}
    \caption{Classification accuracy as a function of different classes. Panel (\textbf{a}) shows  the difference ($Delta$) of class-wise accuracy between 4--6 year-olds and adults on undistorted images (averaged over the salt-and-pepper noise and eidolon experiment). For example while adults recognised 96.66\% of all undistorted \texttt{clock} images, 4--6 year-olds only recognised 36.36\% correctly---resulting in a $Delta$ of 60.30\%. Panel (\textbf{b}) shows confusion matrices for 4--6 year-olds and adults for undistorted images (Noise-level = 0.0) and heavily distorted images (noise level = 0.35) in the salt-and-pepper noise experiment. Rows show classification decision of observers and columns show the ground truth label of the presented category. Transparency of single squares within a matrix represent response probabilities (fully transparent = 0\%, solid red = 100\%). Entries along the negative diagonal represent correct responses; entries which are off the negative diagonal indicate errors.}
    \label{fig:delta_confusion}
\end{figure}

As shown above, already 4--6 year-olds display remarkable levels of object recognition robustness. However, their overall accuracies, even in the noise-free case, are substantially lower than those of older children and adults. Therefore, we ask: Is this difference either due to a general weaker ability to recognise objects---which indicates a qualitative change in object processing and robustness---or could it be due to a weak performance on a subset of categories, which in turn would indicate only a quantitative change in terms of the number of categories they have already acquired? In Figure~\ref{subfig:delta}, we take a closer look at accuracies across different classes. We observe highly non-uniform accuracies: for some classes like ``airplane'', ``car'' etc., 4--6 year-olds have nearly adult-level accuracies. On the other hand, there are also a number classes where young children perform substantially worse, such as ``clock'' or ``knive''. This overall pattern is confirmed when looking at confusion matrices (Figure~\ref{subfig:confusion}), which shows that 4--6 year-olds maintain high performance on a number of classes even for severe levels of noise (as indicated by high accuracies (red-ish entries) on the diagonal).\footnote{Additional confusion matrices can be found in the Appendix~\hyperref[app:confusion]{A.3}.} This may indicate that young children's weaker overall performance is not due to to a generally weaker ability to recognise objects but rather to weak performance on a subset of categories. Even though 4--6 year-olds have not yet acquired robust representations for the same number of categories as adults, they appear almost adult-like regarding some age-appropriate categories they have already acquired. This suggests that the change in robustness along the developmental trajectory is rather quantitative (incremental) and not qualitative in nature.

\subsection*{Back-of-the-envelope calculation: human robustness doesn't require seeing billions of images during lifetime}
\label{sec:envelope_calculation}

So far we have seen that robust object recognition emerges early in development and is largely in place by the age of five. After the age of nine, OOD robustness does not seem to increase substantially. This indirectly indicates that for humans---quite different than for DNNS---more ``data'' (or experience) does not necessarily imply better robustness. As an attempt to quantify this more directly, we estimate the number of ``images'' that young children are exposed to during their lifetime and compare different age groups with different models.

We estimated the amount of external visual input by calculating the total number of fixations during lifetime for each age group. In order to do so we made two assumptions: (a) accumulated wake time for any given age group and (b) fixations per second for any given age group. Calculating the former is straightforward: During development, wake time gradually increases as a function of age. For example 0--1 year-olds are, on average, awake for 11.5 hours a day, whereas adults are awake for 16.5 hours \parencite{thorleifsdottir2002sleep}. We took the mean age of each tested age group and calculated the total accumulated wake time for this particular age in seconds. Estimating the number of fixations per second is more difficult: Fixation duration varies to a great extent (100--2000 ms; e.g., see \cite{young1975survey,karsh2021looking})  and is heavily dependent on age and the given visual task \parencite{galley2015fixation}. Thus, as a reference, we chose a task which is close to an everyday natural setting (a picture inspection task) and for which developmental data is available \parencite{galley2015fixation}. We then calculated fixations per second for each age group based on the fixation duration measured for the mean age of this particular age group. Because there is no available data for adults in the picture inspection task, we estimated the fixation duration of adults by fitting a linear regression line. Fixations per second calculated in this way ranged from 2.56 for 4--6 year-olds to 3.42 for adults (mean adult fixation time of 292 msec; see Table Table~\ref{tab:back_envelope} in the Appendix~\hyperref[app:back_envelope]{A.4} for details).

\begin{figure}[!ht]
     \begin{center}
     \begin{subfigure}[b]{0.75\textwidth}
         \centering
         \includegraphics[width=\textwidth]{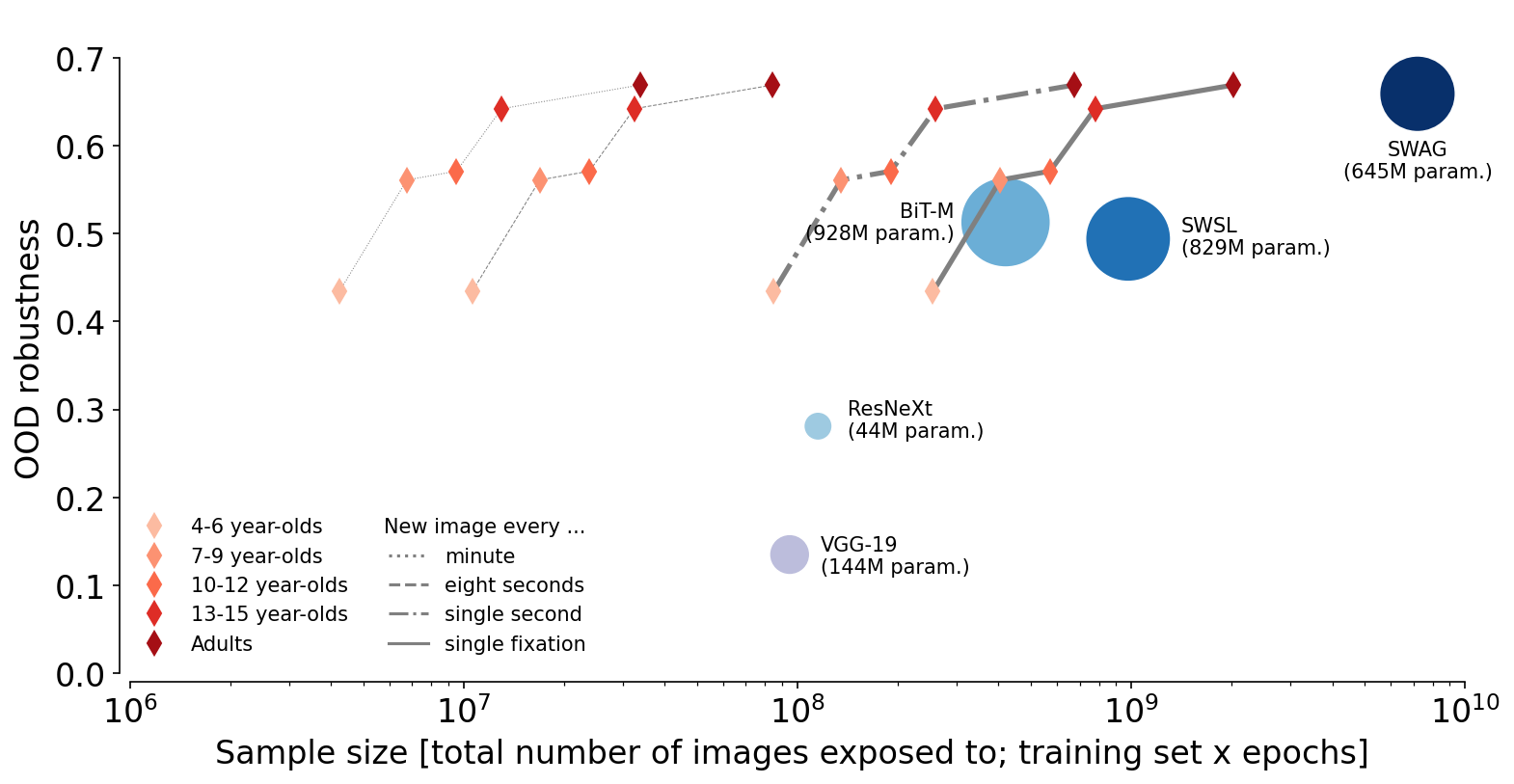}
         \caption{}
         \label{subfig:samplesize}
     \end{subfigure}
     \hfill
     \begin{subfigure}[b]{0.75\textwidth}
         \centering
         \includegraphics[width=\textwidth]{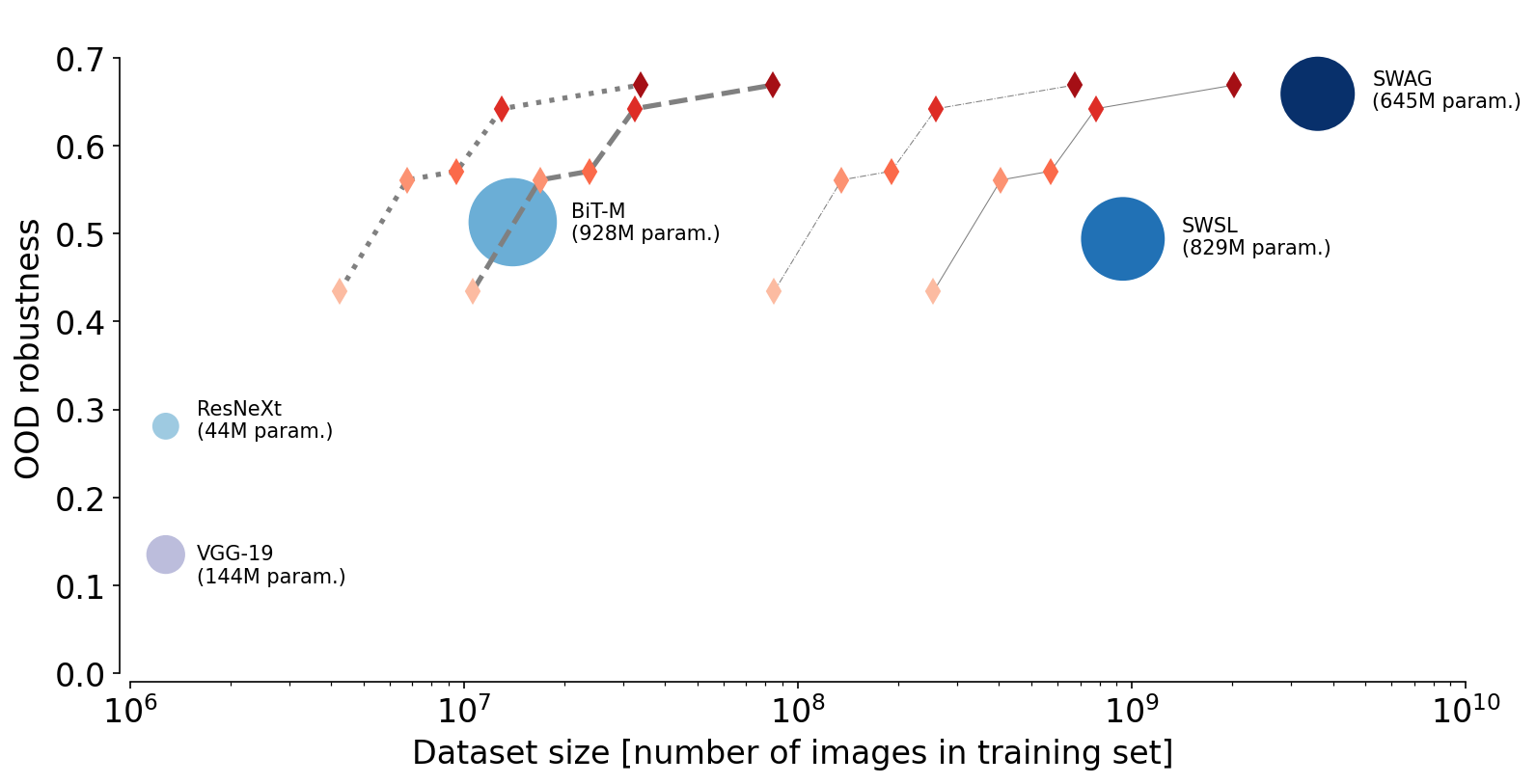}
         \caption{}
         \label{subfig:datasetsize}
     \end{subfigure}
     \hfill
    \end{center}
    \caption{Mean OOD robustness for different age groups and models as a function of \textbf{(a)} sample size and \textbf{(b)} dataset size on semi-logarithmic coordinates. For human observers four different estimates of the amount of visual input are given (indicated by different line types) which results in four different trajectories. We suggest that for the comparison regarding sample size the two most right trajectories, and regarding dataset size the two most left trajectories should be considered (bold lines). Circle area for models reflect number of parameters optimised during training.}
    \label{fig:dataset_samplesize}
\end{figure}

However, given that for extended periods of time in everyday life visual input does not change significantly, one may not want to count each fixation as a new input image. Furthermore, using head-mounted cameras it has been shown that frequency distributions of objects in toddlers' input data are extremely right skewed \parencite{smithDevelopingInfantCreates2018}: Toddlers have only experience with very few objects of a certain category but see those objects (images) very often. It is not clear whether this is just a non-optimal consequence of the natural learning environment of humans or whether the existence of many similar views of the same object plays an important role in object name learning and thus in learning robust visual representations \parencite{clerkin2017real}. To account for these ambiguities, we provide four different estimates regarding the amount of human visual input. As a minimum we assume a new image every minute, whereas as maximum we assume a new image every single fixation. Additionally, and less extreme, we propose a lower (new image every eight seconds) and an upper (new image every single second) estimate between the bounds set by every minute and every fixation.\footnote{Whereas we argue that the maximum---given by a new image every single fixation---constitutes a naturally occurring constraint, the other three estimates can be considered as ``reasonable'' guesses.} Furthermore, there are different choices for counting input images for DNNs. Should every encountered image (\emph{sample size} $=$ training dataset size $\times$ number of epochs, i.e. iterations over the entire training dataset) or the \emph{dataset size} (number of images in training dataset) be considered as visual input? It is unlikely that training on a smaller datasets for an increased number of epochs yields the same increase in robustness as training on a larger datasets. In fact, the evaluated models vary substantially regarding dataset size (1.28M to 3.6B) and epochs (2 to 90). Thus, we decided to plot human data against both metrics, sample size as well as dataset size (see Table~\ref{tab:model_details} in the Appendix~\hyperref[app:back_envelope]{A.4} for details on the calculation of input images for DNNs). Figure~\ref{fig:dataset_samplesize} shows the comparison between different age groups and models regarding OOD robustness and number of input images for DNNs' sample size (Subfigure~\ref{subfig:samplesize}) and dataset size (Subfigure~\ref{subfig:datasetsize}). As a unified measurement of classification robustness we calculated the mean classification accuracy over all moderately and heavily distorted images (salt-and-pepper noise: noise level 0.2 and 0.35, eidolon: reach level 8 and 16) for each age group and all models.

Overall, human object recognition robustness is more data-efficient compared to DNN robustness, irrespective of the choice of metric. For example, focusing on sample size, we find that the two least data-hungry models (ResNeXt and VGG-19) have been exposed to about as much input as 4--6 year-olds (notably only looking at the two highest of our image number estimates) but are 15--30\% less robust. The only model which is comparable to 13--15 year-olds in terms of OOD robustness as a function of sample size is SWAG (0.642 vs. 0.659). However, even when counting all fixations as input images---most likely an overestimation of the human external visual input---SWAG needs about ten times more data to achieve human-like OOD robustness (779M vs. 7.2B). Probably a more plausible comparison is accomplished by looking at the two more moderate estimates---a new image every single second (dashdotted line) or every eight seconds (dashed line)---and comparing them with sample size or dataset size, respectively. Regarding sample size we find that all three models achieving high OOD robustness (BiT-M, SWSL, and SWAG) need substantially more data than humans to do so. The same is true if we consider datset size, except for BiT-M which aligns with the human OOD trajectory (similar OOD robustness of a 6--7 year-old and similar dataset size of a 6--7 year-old if every awake second is equated with a new image).

It may be important to recall that except for VGG-19, all of the investigated models were chosen since they were the \emph{most robust} ResNet-based models for a given training dataset size according to the model-vs-human bechmark \parencite{geirhos2021partial}. Thus these models represent some of the current best models in terms of data-efficient robustness---comparisons with the many other DNNs would have resulted in even larger discrepancies between humans and DNNs.

Looking only at the different DNNs we find that BiT-M achieves similar robustness to SWSL with a much smaller dataset (14M vs. 940M). However, this gap almost vanishes when looking at sample size. This indicates that the total number of images exposed to during training (sample size) seems to matter more in terms of OOD robustness than plain dataset size. This may be the case since due to data augmentation, images are not exactly the same for every epoch. It has been shown that common data augmentations (such as random crop with flip and resize, colour distortion, and Gaussian blur) lead to higher OOD robustness \parencite{perez2017effectiveness, shorten2019survey, mikolajczyk2018data}. Regarding the number of parameters optimised during training---the area of the circles in the figure---we do not find any direct link to OOD robustness.  

\subsection*{Different strategies: big models aren't like children, but children are like small adults}
\label{sec:strategy_development}
In the sections above we have seen comparisons of overall accuracy and robustness and how this is related to the amount of visual input. While accuracy increases with age (i.e., older children successively categorise more and more images correctly), it is still unclear whether children just gradually acquire more categories, or whether they go through a more radical change of perceptual strategy at some point (at least in terms of overall behaviour, teenagers clearly change a lot). To this end we performed two analyses aimed at understanding how object recognition \emph{strategies} change (if at all) during childhood and adolescence. The first analysis is related to the image cues used for object recognition (shape or texture), and the second related to image-level errors (error consistency).

\subsubsection*{Texture-shape cue-conflict: no evidence of a strategy change}
\label{sec:cueconflict}

\begin{figure}[!ht] 
    \begin{center}
    \includegraphics[scale=.8]{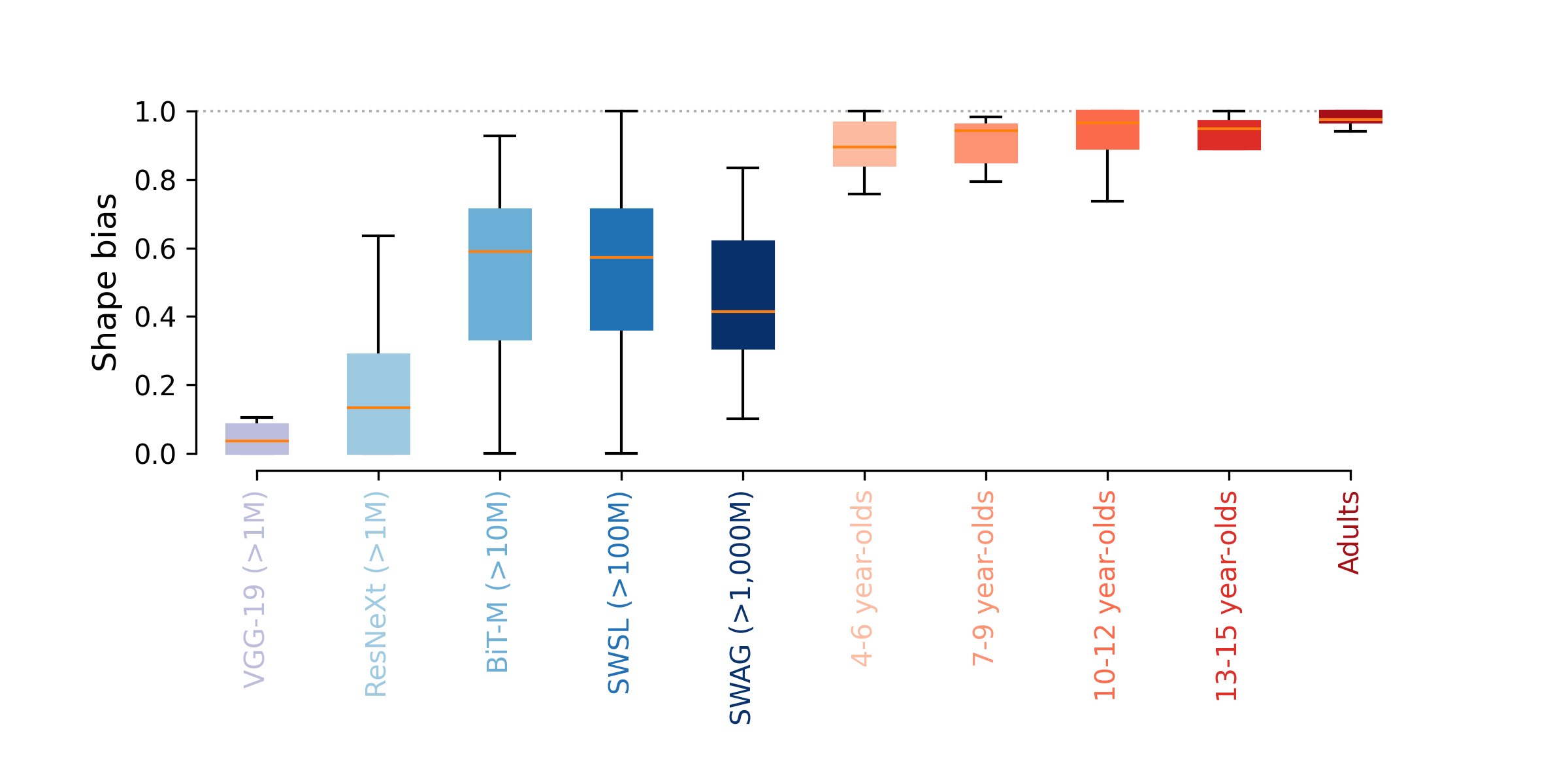}
    \end{center}
    \caption{Shape vs. texture biases of different models and age groups. Box plots show \textit{category-dependent} distribution of shape/texture biases (shape bias: high values, texture bias: low values). For example, 4--6 year-olds show a shape bias of 0.88, meaning that of all correct responses they decided in 88\% of cases based on shape- and in 12\% of cases based on texture-cues. The dotted line indicates the maximum possible shape bias (100\% shape based decisions). Shape vs. texture biases for individual categories are shown in Figure~\ref{fig:shape_bias} in the \hyperref[sec:appendix]{Appendix}.}
    \label{fig:shape_bias_box}
\end{figure}

\textcite{geirhos2019imagenet} and \textcite{baker2018deep} have shown that adults and ImageNet-trained DNNs have a clear discrepancy in object recognition strategy. While human adults base their classification decisions on object shape, DNNs are much more prone to using texture cues instead. In order to determine whether children are more similar to adults or to DNNs in this regard, we evaluated performance on texture-shape cue-conflict images. These images contain conflicting shape and texture information (e.g., the shape of a cat combined with the texture of an elephant; example stimuli shown in Figure~\ref{fig:cue-conflict}). It may be worth pointing out that there is no right or wrong answer in those cases---both the correct texture category or the correct shape category is considered as a correct response. Instead, we are interested in understanding whether decisions are consistent with the shape or the texture category. The results are visualised in Figure~\ref{fig:shape_bias_box}. The exact fractions of shape vs. texture biases and the category wise proportions of texture vs.\ shape decisions are shown in the Appendix \hyperref[app:shape_bias]{A.5}. Those results clearly show that irrespective of age, humans have a very strong shape bias (between approximately 0.88 and 0.97), and there is no evidence to suggest any change of strategy during human development in this regard.\footnote{Note that we use the term ``shape bias'' for the tendency to use object shape as the crucial feature to identify an object. However, in the developmental literature the term ``shape bias'' is used to describe the tendency of young children to use object shape as the crucial property to generalize names to objects that were not seen before.} Even models trained on extremely large datasets, however, still don't have a shape bias comparable to humans. In other words, when it comes to using texture or using shape, big models are not like children, but children are like small adults.

\subsubsection*{Error consistency: distorted input serves as a magnifying glass for object recognition strategies}
\label{sec:error}

\begin{figure}[ht]
    \begin{center}
    \includegraphics[scale=.43]{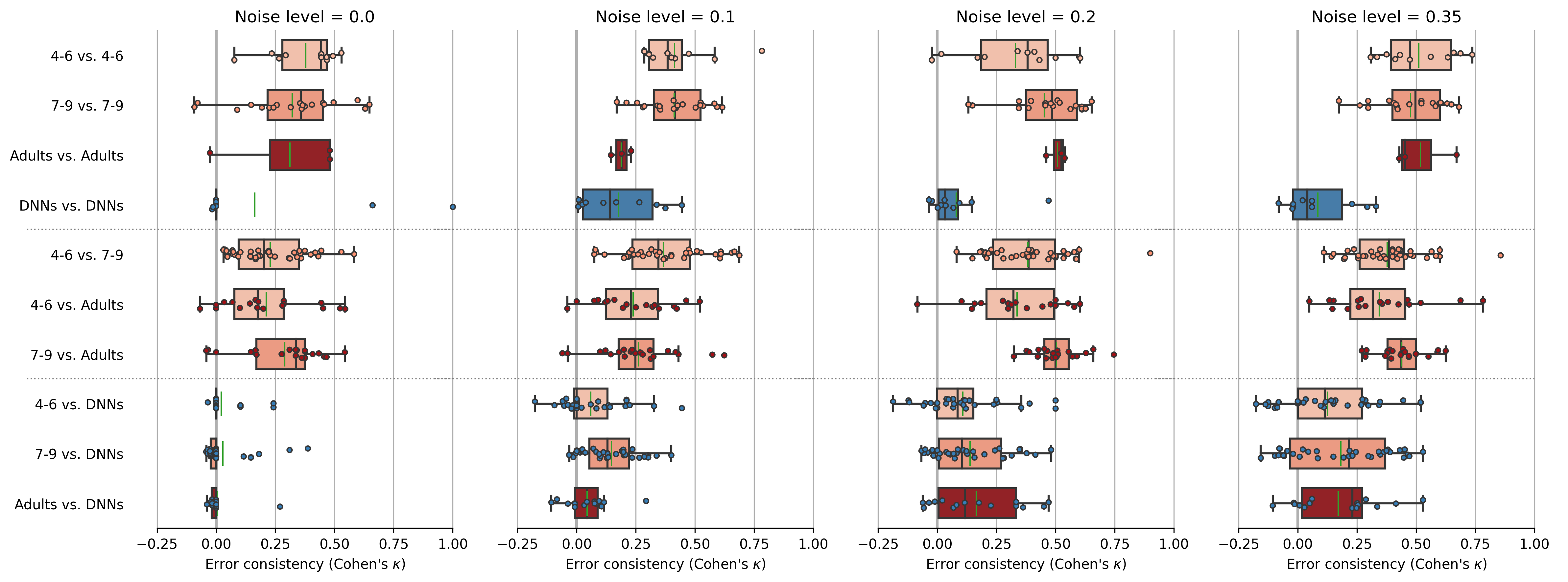}
    \end{center}
    \caption{Distorted input serves as a magnifying glass for object recognition strategies---irrespective of age, children make errors on the same noisy images as adults; at the same time, models make errors on different images as humans. The plot shows error consistency as measured by Cohen's kappa ($\kappa$) for different distortion levels split by different within- and between-group comparisons (\textcolor{4-6}{4--6}, \textcolor{7-9}{7--9}, \textcolor{adults}{adults}, and \textcolor{midblue_errorK}{DNNs}) for the salt-and-pepper noise experiment. $\kappa = 0$ indicates chance level consistency (i.e., both systems are using independently different strategies), $\kappa > 0$ means consistency above chance level (i.e., both systems are using similar strategies) and $\kappa < 0$ means inconsistency beyond chance level (i.e., both systems use inverse strategies). Plots are horizontally divided in three subsection*s: Upper subsection* (within-group comparisons), middle subsection* (between-group comparisons humans only), and lower subsection* (between-group comparison humans and DNNs). Colored dots represent error consistency between two single observers  (one from each group featured in the comparison). Box plots represent the distribution of error consistencies from observers of the two given groups. Boxes indicate the interquartile range ($IQR$) from the first ($Q1$) to the third quartile ($Q3$). Whiskers represent the range from $Q1-IQR$ to $Q3+IQR$. While vertical black markers indicate distribution medians, faint vertical green markers indicate distribution means.}
    \label{fig:error_consistency}    
\end{figure}

In the previous section, we have seen that there does not appear to be a radical change of perceptual strategy during childhood when it comes to using texture or shape cues to identify object categories. Nonetheless, all of the previously analysed measures are fairly coarse. Both accuracy and shape bias analyses could potentially overlook more subtle changes of strategy---if, for example, adults struggle with certain images that children find easy, and vice versa, then we would end up with very similar aggregated decisions (as measured by accuracy) despite highly different image-level decisions. Therefore, we here looked at image-level decision patterns through the lens of error consistency \parencite{geirhos2020beyond}. Error consistency is a quantitative analysis for measuring whether two decision-makers systematically make errors on the exact same stimuli. Error consistency between two individual observers is calculated in three steps: First, \textit{observed error overlap} is calculated by dividing the total number of \textit{equal} responses---in which both observers either classified an image correctly or incorrectly---by the total number of images both observers have evaluated. Second, since even two completely independent observers with high accuracy will necessarily agree on many trials by chance alone, \textit{error overlap expected by chance} is calculated (based on the assumption of binomial observers). Third, the empirically observed error overlap is compared against the error overlap expected by chance via Cohen's $\kappa$, which quantifies the agreement of two observers considering the possibility of the agreement occurring by chance.

Here, we compared four different decision-makers (4--6 year-olds, 7--9 year-olds, adults and DNNs) using error consistency. We performed all possible within-group (e.g., 4--6 year-olds with 4--6 year-olds) and between-group (e.g., 4--6 year-olds and DNNs) comparisons.\footnote{Since not all children responded to all stimuli (see \hyperref[sec:general]{method} section), error consistency sometimes was only calculated on a subset of stimuli. In those cases, we set a minimum constraint of 20 individual stimuli, which had to be evaluated by both observers. Otherwise error consistency was not calculated.} In Figure~\ref{fig:error_consistency} error consistency is visualised for all difficulty levels of the salt-and-pepper-noise experiment split by different within- and between-group comparisons; the (similar) error consistency plot for the eidolon experiment can be found in Figure \ref{fig:errorK_eidolon} in the Appendix~\hyperref[app:shape_bias]{A.6}.

It may be worth pointing out the central patterns: First of all, in line with \parencite{geirhos2020beyond, geirhos2021partial} human-to-human consistency is generally high, and while it is highest within the same age group it is also well beyond chance in all between-age-group comparisons. Second, human-to-human error consistency \emph{increases} as the tasks becomes harder (i.e.\ with increasing noise level). Third, regarding comparisons involving DNN models, a very different pattern emerged. Model-to-model consistency starts at chance level and does not increase substantially beyond chance as a function of noise level (highest at noise level 0.1; \textit{mean} $=$ 0.18). Furthermore, regardless of age, model-to-human consistency is at chance level for noise level zero. And also for distorted images, model-to-human consistency (mean over all model-to-human comparisons for distorted images $=$ 0.127) is far below human-to-human consistency (mean over all human-to-human between-age-group comparisons for distorted images $=$ 0.361). Thus, it almost appears as though distorted input serves as a magnifying glass for object recognition strategies---irrespective of age, children make errors on the same noisy images as adults; at the same time, models make errors on different images as humans.

\section*{Discussion}
\label{sec:dis}

We investigated the developmental trajectory of core object recognition robustness to asses whether human OOD robustness is a result of training (experience) on a very large amount of visual input--similar as in currently OOD-robust DNNs. To this end we collected 23,474 psychophysical trials from 146 children and nine adults and compared their OOD performance against five DNNs trained on datasets of different sizes. To our knowledge this is the first study to directly compare children, adolescents, and different DNNs in a psychophysical core object recognition task using an experimental protocol also employed for adults and in machine learning.\footnote{There is one unpublished investigation comparing children and DNNs that was presented at the 20th Annual Meeting of the Vision Sciences Society \parencite{ayzenberg2020young}. They find that young children (4--5 year-olds) display remarkable object recognition abilities and outperform a VGG-19 and a ResNet-101 model on perturbed images. Thus, their findings fit our results regarding classification accuracy. However, the current study extends their findings by covering a larger age range, using naturalistic images with parameterized distortions and more response categories. Furthermore, we provide a range of additional analysis such as accuracy delta between children and adults, confusion matrices, human-model comparison regarding OOD robustness and input images, texture-shape cue-conflict analysis, and error consistency. Additionally, we compare human data with some of the most powerful models to date. \label{foot:ayzenberg}}

We find that, first, human OOD robustness develops very early and is essentially in place by the age of five. While there may be a small increase in robustness as function of age, by the time children reach middle childhood they have approximately obtained adult-level robustness (see Figure~\ref{fig:accuracy}, right column, normalized accuracy). This finding fits with neuroscience data showing that brain maturation relevant for object recognition reaches adult-level at this point of development \parencite{golarai2010differential, scherf2007visual, conner2004retinotopic, ben2007contrast}. Furthermore, we find that young children did not perform uniformly weaker on all categories (see Figure~\ref{fig:delta_confusion}), indicating that the observed overall improvement in accuracy is due to acquisition of new categories rather than to a global change in representation and information processing (see Figure~\ref{fig:accuracy}, left column, accuracy). This allows even 4--6 year-olds to outperform DNNs trained on standard ImageNet. Second, by estimating the visual input for human observers at different points during development, we find that---in contrast to current DNNs---human OOD robubstness requires relatively little external visual input (see Figure~\ref{fig:dataset_samplesize}). This indicates that in humans, OOD robustness may not be achieved solely by sheer quantity of training data alone. Third, the former two findings are supported by our observations that all tested age groups employ similar object recognition strategies as indicated by a similar shape-bias (see Figure~\ref{fig:shape_bias_box}) and high error consistency across different age groups and difficulty levels (see Figure~\ref{fig:error_consistency}).

Taken together, these findings suggest that for both humans and DNNs, robust visual object recognition is possible, but achieved by different means. While human robustness seems fairly data-efficient, machine robustness, at least today, is data-hungry.\footnote{Assuming that (a) from a evolutionary point of view robustness is a crucial feature of our visual system (e.g., food detection and predator avoidance under difficult conditions such as in the dark or during snow, rain or fog), and that (b) it is not feasible to gather enough visual input during development (even when counting all fixations as new images humans are not gathering as much images as robust DNNs are trained on) to achieve robustness by the same means as modern DNNs (large-scale training), there is a high selection pressure for highly data-efficient learning. In other words, from an evolutionary point of view, our back-of-the-envelope calculation shows that human data-efficient robustness should not be surprising.} In other words, there are two different systems with the same property which came about in different ways---a phenomenon which in biology is called \emph{convergence} \parencite{mcghee2011convergent}. As an example, consider the ability to fly, which emerged at least three different times during evolution: in mammals (e.g., bats), in sauropsida (e.g., birds) and in insects (e.g., dragonflies). \textcite{lonnqvist2021comparative} recently argued that considering DNNs and humans as different visual species, and adopting an approach of comparative biology by focusing on the differences rather than the similarities, is a promising way to understand visual object recognition. Accordingly, in what follows, we elaborate on possible differences between human and DNN vision, which might explain the difference in data-efficiency to solve robust object recognition.\footnote{This elaboration is not thought to be exhaustive, but rather to address aspects in which the present study is limited and which we suggest are promising for future research.}

First, there might be a difference in \emph{data-quality}, which allows humans to form more robust robust representation from limited data. While human data is continuous and egocentric \parencite{bambach2017egocentric}, this is not the case for standard image databases. Recent advances in data collection using head-mounted cameras allow for developmentally realistic first-person video datasets \parencite{sullivan2020saycam, fausey2016faces, jayaraman2015faces, bambach2018toddler}. While studies have shown that models trained with such biologically plausible datastreams form powerful, high-level representations \parencite{orhan2020self} and achieve good neural predictivity on different areas across the ventral visual stream \parencite{zhuang2021unsupervised}, others find that in order to match human performance in object recognition tasks models would need millions of years of natural visual experience \parencite{orhan2021much}. A further difference regarding the data-quality of humans vs. machines lies in the modality of the data; while model input is most often unimodal, human input is multi-modal. It has been shown that the availability of information across different sensory systems is linked to the robustness of human perception (e.g., see \cite{ernst2004merging,gick2009aero,von2012multisensory, sumby1954visual}). Regarding vision, \textcite{berkeley1709essay} famously argued that ``touch educates vision''. Affirmatively, a recent study demonstrated that neural networks trained in a visual-haptic environment (compared to networks trained on visual data only) form representation that are less sensitive to identity-preserving transformation such as variations in view-point and orientation \parencite{jacobs2019can}. Taken together, the continuous, egocentric and multi-modal nature of human training data might explain why current DNNs are not as data-efficient as humans. Accordingly, a limitation of our study is the lack of systematic variation in the quality of training data. Perhaps providing DNNs with high quality training data even current DNN architectures could achieve OOD robustness with as little data as humans. Thus, future research should systematically acquire multi-modal datasets of varying quality and evaluate the trained models on OOD datasets. 

Second, humans may rely on different \emph{inductive biases}---i.e. constraints or assumption prior to training (learning)---allowing for more data-efficient learning. Especially intuitive theories, e.g., intuitive physics, theory of mind or implicit knowledge about the causal structure of the world, might lead to efficient processing of the available data (e.g., see \cite{lake2017building, marcus2020next} or \cite{goyal2020inductive}) for the role of inductive biases in OOD robustness in general). For example, once learned that the representation of a certain object does change based on certain physical conditions (such as lighting or distance), intuitively knowing that all objects obey the laws of physics and behave in a causally predictable way, should facilitate object recognition for other objects which are affected in similar ways. Human inductive biases are the product of million years of evolution and are built in right from the start (birth). Thus in order to further disentangle the influence of evolution vs.\ lifetime experience it would be interesting to investigate the developmental trajectory during infancy. In this regard, the present study is limited, however, because the employed experimental set-up does not allow to test children younger than four years of age.

Third, an exciting possibility is that humans enlarge their initial dataset provided through external input by creatively using already encountered instances to create new instances during \emph{offline states}---a concept similar to what in reinforcement learning is called \emph{experience replay} (e.g., see \cite{o2010play,lin1991programming,lin1992self,mnih2015human}). The idea is that during imagination and dreaming, stored memories are combined to generate new training data (e.g., see \cite{deperrois2022learning}). Thus, additionally to the external input provided by the sensory system, an internal generative model provides the visual system with additional training data. Putting this into context, one could argue that humans and DNNs might be similar to the extent that they both rely on large-scale datasets to solve object recognition robustness, but are, however, very different in how they attain such large datasets: While DNNs are entirely dependent on external input, humans are somewhat self-sufficient by producing their own training data from limited external input. Assuming that this hypothesis about the emergence of human OOD robustness is true, the question is whether learning during offline states could make DNNs as data-efficient as humans. In the present study, we only compare how much \emph{external} input is required to achieve high OOD robustness. Our results are thus not suited to answer this question. However, recently, \textcite{deperrois2022learning} proposed a model based on generative adversarial networks (GANs), which captures the idea of learning offline states by distinguishing between wake-states where external input is processed, and offline states where the model is trained by a generative model either by reconstructing perturbed images based on latent representations (similar to simple memory recall as during non-REM sleep) or by generating new visual sensory input based on convex combinations of multiple randomly chosen stored latent representations (similar to the rearranging of stored episodic patterns during REM sleep). Experiments with these models show that introducing such offline states increases robustness and the near separability of latent representations. Further evidence for the benefit of learning during offline states comes from world-model-based reinforcement learning. It has been shown that reinforcement learning agents are able to solve different task only being trained in a latent space which could be conceptually associated with the model's dreams, imagination or hallucinations (e.g., see \cite{zhu2020bridging, hafner2019dream, ha2018world}).

The three above described differences between humans and DNNs might explain the difference in data-efficiency found in the present study. However, they are arguably only a small subset of all differences which might account for the differences in data-efficiency. What is clear, however, is that object recognition robustness is not only solvable by a single approach. In evolution, there are often many paths to the same feature. Why a certain path was taken by a given system can only be understood by looking at the environmental constraints present during phylogenesis. Not being biological systems, DNNs were not exposed to similar evolutionary pressure as humans and thus data-efficiency seems not to be as important as for humans. Accordingly, it comes as no surprise that DNNs are less data-efficient than humans. However, the data-efficiency of humans seems to be a crucial feature of the human visual system. Thus, in order to truly understand the robustness of human vision, we need not only to model the behaviour (OOD robustness) but also the means by which it is achieved.

\section*{Conclusion}
\label{sec:conclusion}

Recent improvements in OOD robustness in machine learning are primarily driven by ever-increasing large datasets, with models trained on several billions of images. However, humans achieve remarkable OOD robustness at a very early point in life. Our investigations and calculations suggest that children learn a lot from relatively little data. Children benefit from experience, but they do not require the same amount of experience as state-of-the-art neural network models, indicating that they are achieving OOD robustness by different means as DNNs. It seems to be the case that even the entire time children have already lived is not long enough to gather as much input as current data-rich DNNs. The human visual system appears highly data-efficient, which may be an evolutionary advantage. It remains an open question what the sources of this data-efficiency are: Is it due to the accumulation of high-quality data alone? High-quality data combined with suitable inductive biases and mechanisms to ``upcycle'' data in order to enlarge the training dataset during offline states such as dreaming? We believe that comparing children with adults and DNNs is a fruitful approach to understand data-efficiency in humans and to perhaps inspire a healthy diet for current data-hungry models, without sacrificing their robustness.

\clearpage

\section*{Acknowledgements}
We would like to thank the members of the Wichmann-lab for their support and insightful discussions. Special thanks goes to Uli Wannek for excellent technical advice and Silke Gramer for extremely kind and patient administrative support. Additionally, we would like to express our gratitude to Gert Westermann and Hannes Rakoczy for their advice and help with designing the children's study and to all children and teachers who participated in our study. Felix Wichmann is a member of the Machine Learning Cluster of Excellence, funded by the Deutsche Forschungsgemeinschaft (DFG, German Research Foundation) under Germany’s Excellence Strategy – EXC number 2064/1 – Project number 390727645. Some preliminary parts of this work have been presented as an oral at the Annual Meeting of the Vision Sciences Society 2021 and at the 3rd Workshop on Shared Visual Representations in Human and Machine Intelligence (SVRHM 2021) of the Neural Information Processing Systems (NeurIPS) conference. 

\section*{Author contributions}

The project was initiated and lead by L.S.H. All authors planed and designed the experiments and data analysis. L.S.H. collected the psychophysical data, evaluated the models and analysed the data. The first draft was written by L.S.H. with significant input from R.G. and F.A.W. All authors contributed to the final version of the manuscript. 

\section*{Competing interests}

The authors declare no competing interests.

\section*{Data and Code}
All code and data are available from this repository:\\ \url{https://github.com/wichmann-lab/robustness-development}

\clearpage

\appendix
\section*{Appendix}
\label{sec:appendix}

In this Appendix we provide further experimental details as well as supplementary plots and details about the data analysis. Appendix \hyperref[app:gamification]{A.1} provides some exemplary screenshots of the visual details of the user interface. Details and characteristics of the tested sample of children and adults are reported in Appendix \hyperref[app:demo]{A.2}. A full set of confusion matrices for 4--6 year-olds and adults for both experiments can be found in Appendix \hyperref[app:confusion]{A.3}. Supplementary details regarding the estimation of human input images and dataset and sample size of evaluated models are given in Appendix \hyperref[app:back_envelope]{A.4}. Further, we provide additional details about the results of the cue-conflict experiment in Appendix \hyperref[app:shape_bias]{A.5}. Finally, the error consistency plot of the eidolon experiment can be found in Appendix \hyperref[app:errorK]{A.6}.

\subsection*{A.1 Gamification}
\label{app:gamification}

\begin{figure}[!ht]
     \centering
     \captionsetup{justification=centering}
     \begin{subfigure}[b]{0.45\textwidth}
         \centering
         \includegraphics[width=\textwidth]{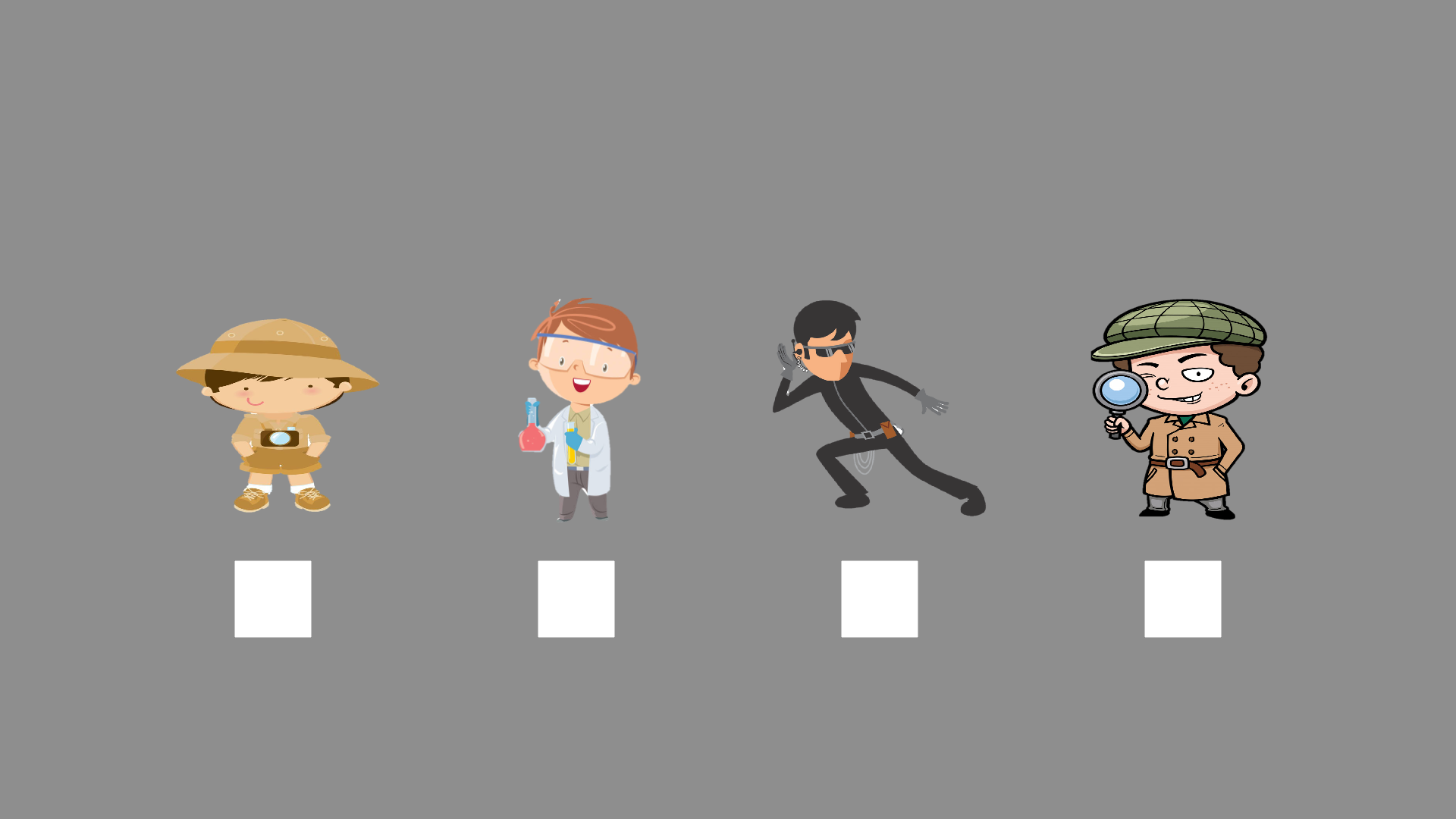}
         \caption{Character selection at the beginning of each session.}
         \label{fig:y equals x}
     \end{subfigure}
     \hfill
     \begin{subfigure}[b]{0.45\textwidth}
         \centering
         \includegraphics[width=\textwidth]{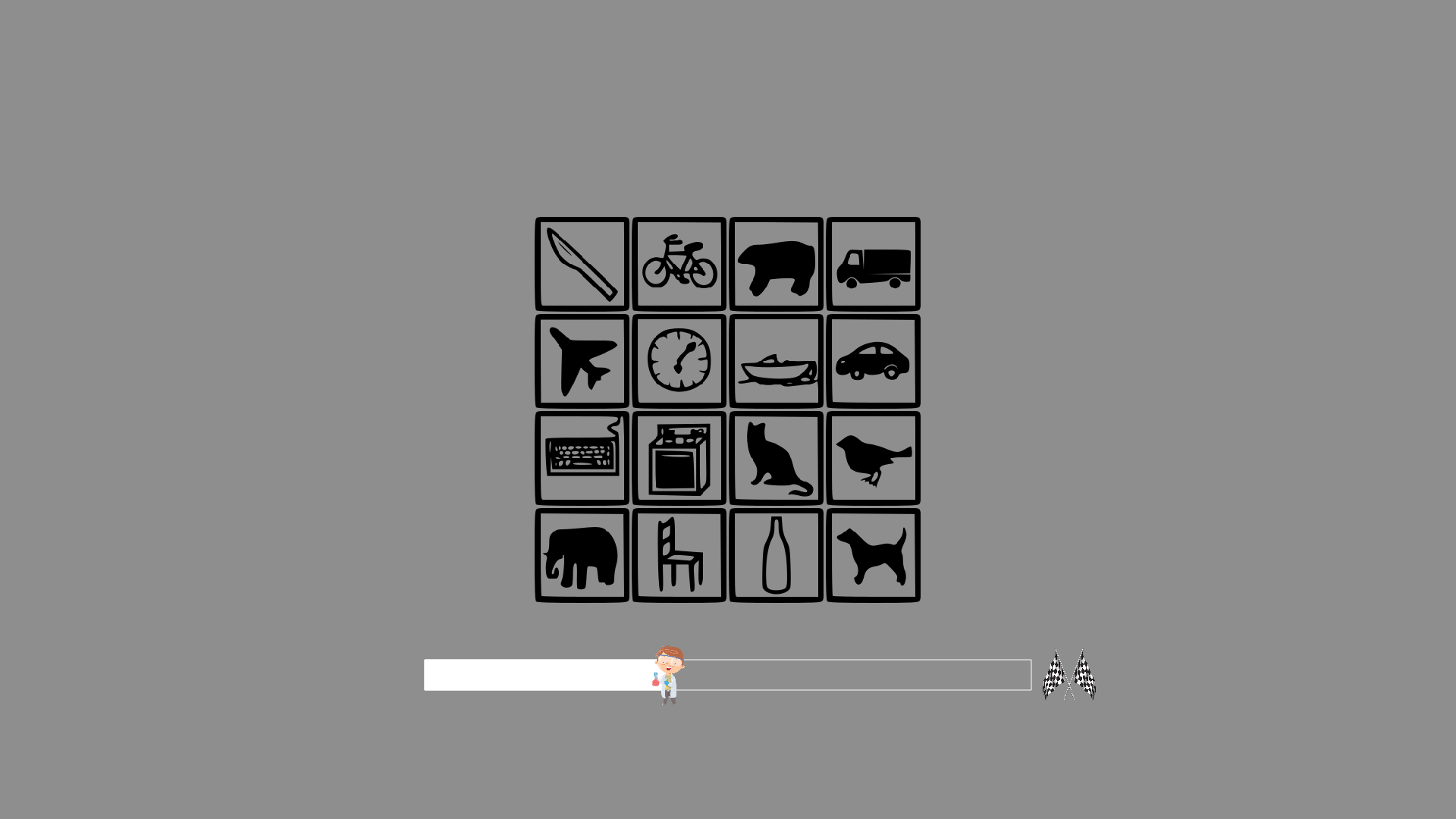}
         \caption{Response screen with gamified progress-bar.}
         \label{fig:three sin x}
     \end{subfigure}
     \hfill
     \vspace{2cm}
     \begin{subfigure}[b]{0.45\textwidth}
         \centering
         \includegraphics[width=\textwidth]{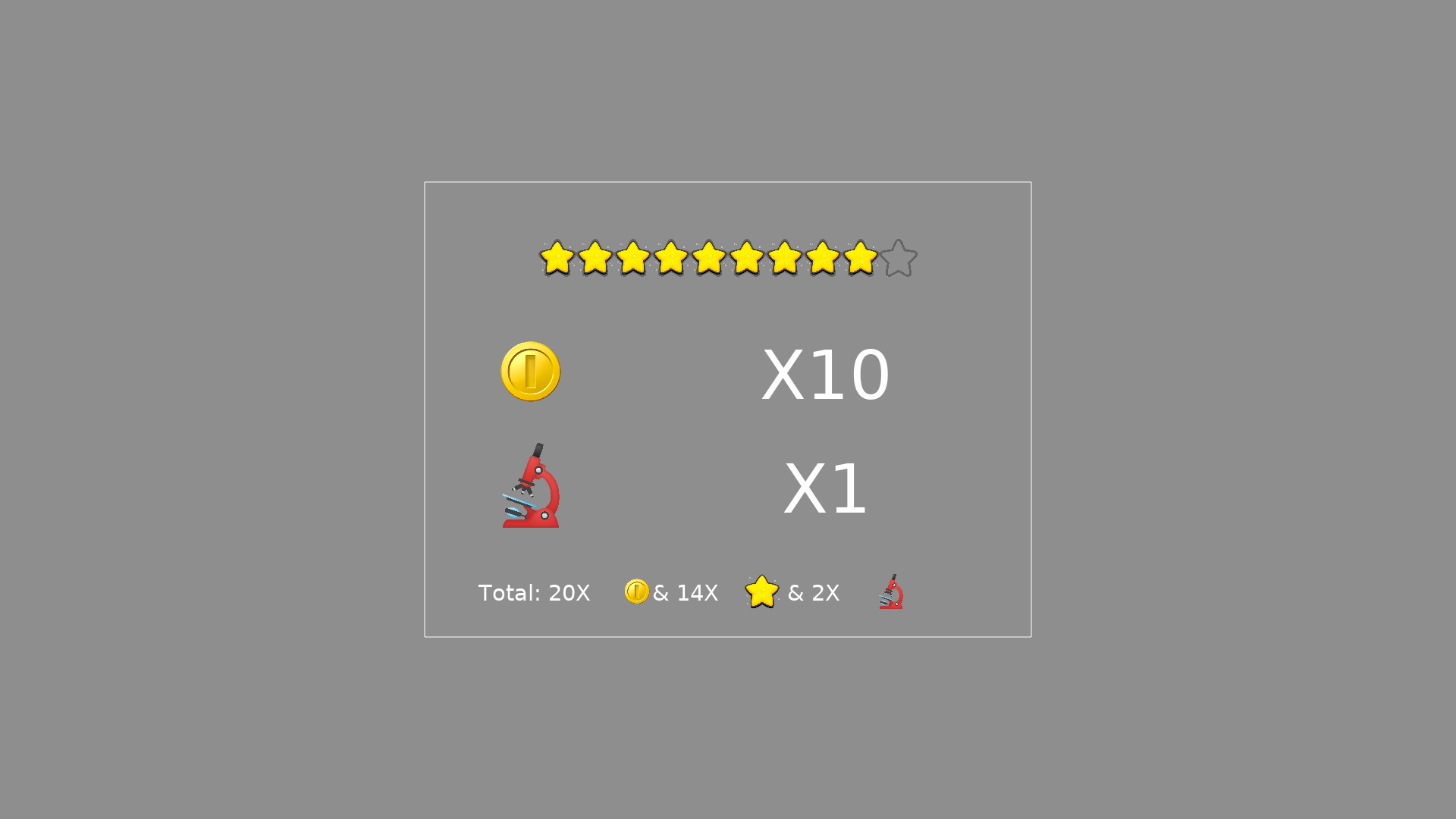}
         \caption{Score display with star- and coin-scores and emblem.}
         \label{fig:five over x}
     \end{subfigure}
     \hfill
     \\
     \caption{Screenshots at different time-points during the experiment.}
        \label{fig:gamification}
\end{figure}
\clearpage

\subsection*{A.2 Demographic characteristics of participants and observations}
\label{app:demo}

\begin{table}[h]
\begin{center}

\caption {Descriptive statistics of participants and observations split by experiments. Sample size and quantity of observations (\textit{n}), as well as mean (\textit{M}) and standard deviation (\textit{SD}) for age and trials within observer groups. Gender distribution (\male/\female) is given in percentages. Note that for adults, trial \textit{M} equals the total number of trials in the respective experiment and trial \textit{SD} is zero because they completed all trials of that particular experiment.}
    \small{\begin{tabular}{@{}llrrrrrrrr@{}}
    \toprule
      &  & &
      \multicolumn{2}{c}{Age within group} &&& \multicolumn{3}{c}{Trials} \\
      \cmidrule{4-5} \cmidrule{8-10} Age group & Experiment &
     \textit{n} & \textit{M} & \textit{SD} & \male/\female && \textit{n} & \textit{M} & \textit{SD} 
     \\
     \midrule
     4--6 year-olds & Noise & 15 & 5.13 & 0.64 & 33/67 && 1240 & 62.66 & 54.96 \\
    & Eidolon & 11 & 5.27 & 0.65 &  64/36 && 1234 & 102.83 & 92.06 \\
    & Cue-conflict & 21 & 5.29 & 0.64 & 62/38 && 1292 & 61.52 & 36.12 \\
    \midrule
     7--9 year-olds&Noise & 9 & 8.11 & 0.78 & 47/53 && 1840 & 204.44 & 112.60\\
    & Eidolon & 14 & 8.36 & 0.93 & 43/57 && 1708 & 127.14 & 61.57\\
     & Cue-conflict &11 & 8.09 & 0.94 & 55/45 && 2020 & 183.63 & 118.60\\
    \midrule
     10--12 years-olds& Noise & 15 & 11 & 0.85 &47/53 && 1880 & 125.33 & 71.9\\
     &Eidolon & 14 & 11.14 & 0.77 & 43/57 && 1820 & 130.00 & 72.64\\
     &Cue-conflict &12 & 11.08 & 0.90 & 50/50 && 2080 & 173.33 & 126.01\\
    \midrule
     13--15 years-olds& Noise & 9 & 14.22 &0.97 & 44/56 && 1280 & 142.22 & 82.12\\
     &Eidolon & 7 & 14.00 & 1.00 & 71/29 && 1700 & 242.86 & 76.10\\
     & Cue-conflict &8 & 14.38 & 0.74 & 50/50 && 2260 & 282.50 & 107.14\\
     \midrule
     Adults & Noise &3 & 28.33 & 5.51 & 33/67 && 960 & 320.00 & 0.00\\
     & Eidolon & 3 & 26.00 & 2.65  & 67/33 && 960 & 320.00 & 0.00\\
     & Cue-conflict & 3 & 29.33 & 2.89 & 33/67 && 1200 & 400.00 & 0.00\\
    \bottomrule
    \label{tab:demo}
    \end{tabular}}
    \end{center}
\end{table}

\newpage

\subsection*{A.3 Confusion matrices}
\label{app:confusion}

\begin{figure}[!ht]
     \centering
     \begin{subfigure}[b]{0.95\textwidth}
         \centering
         \includegraphics[width=\textwidth]{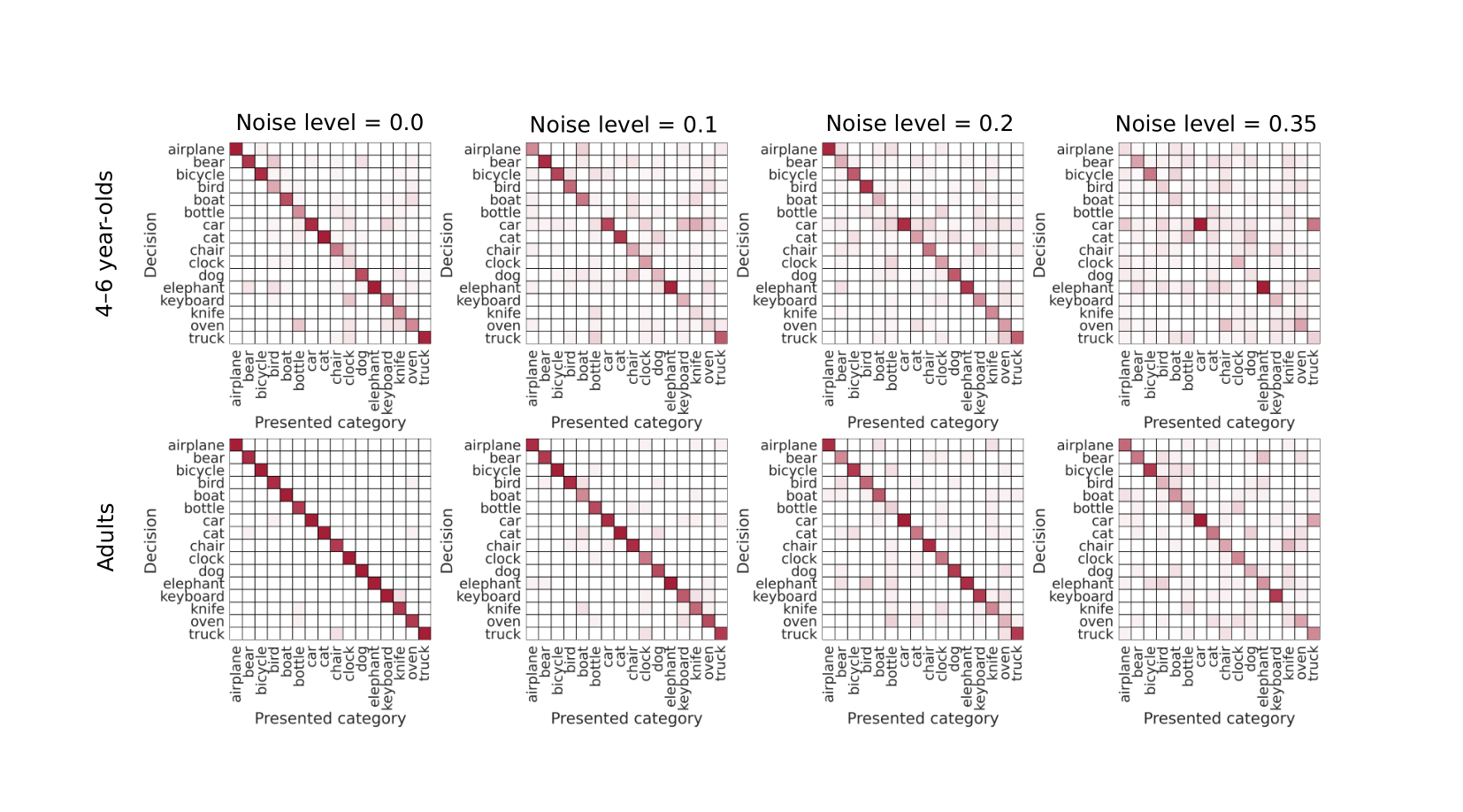}
         \caption{Salt-and-pepper noise}
         \label{fig:y equals x}
     \end{subfigure}
     \hfill
     \begin{subfigure}[b]{0.95\textwidth}
         \centering
         \includegraphics[width=\textwidth]{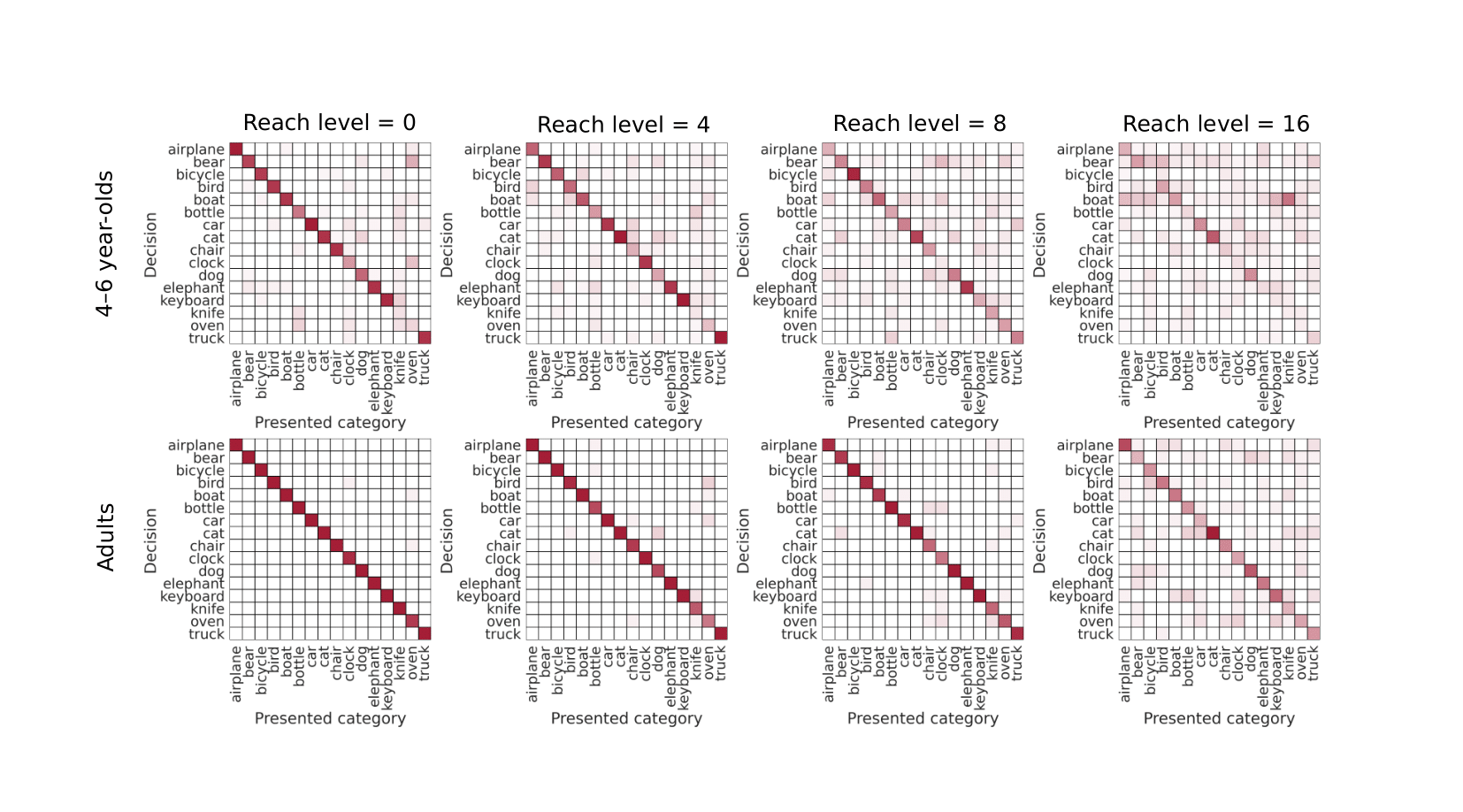}
         \caption{Eidolon}
         \label{fig:three sin x}
     \end{subfigure}
     \hfill
     \\
     \caption{Confusion matrices for 4--6 year-olds and adults across all difficulty levels in the salt-and-pepper noise (\textbf{a}) and the eidolon (\textbf{b}) experiment. Rows show classification decision of observers and DNNs and columns show the ground truth label of the presented category. Transparency of single squares within a matrix represent response probabilities (fully transparent = 0\%, solid red = 100\%). Entries along the negative diagonal represent correct responses; entries which are off the negative diagonal indicate errors.}
        \label{fig:confusion_appendix}
\end{figure}
\clearpage

\subsection*{A.4 Back-of-the-envelope calculation}
\label{app:back_envelope}

\begin{table}[!ht]
\centering
\caption{Details regarding the estimation of the number of input images for human observers. The estimate of accumulated \textit{Wake time} (in millions of seconds) is based on \parencite{thorleifsdottir2002sleep}. \textit{Fixation Duration} (in milliseconds) refers to the fixation duration in a picture inspection task \parencite{galley2015fixation} and is used to calculate \textit{Fixations per second}. Because there was no available data regarding the fixation duration of adults in this task we assumed a linear relationship between age and fixation duration and used the children's data to fit a simple regression model to estimate the fixation duration of adults ($\hat{y} = -4.33X+413.66$). Plugging in the mean age of adults (\textit{M} = 28) yields a predicted fixation duration of 292.42 for adults. \textit{Min} (in millions) refers to the minimal assumed number of input images---a new image every minute. \textit{Max} (in millions) refers to the maximal possible number of input images---a new image every single fixation. \textit{Lower} and \textit{Upper} (in millions) refers to a---what we believe---reasonable estimate of the lower and upper bound of input images encountered during lifetime. The lower bound is calculated by scaling the total number of fixations by 24 (new images approximately every eight seconds) and the upper bound by 3 (new images approximately every single seconds). E.g., at the age of five a child has been awake for approximately 99.41 million seconds; it has made about 254.5 million fixations during this time (99.41 $\times$ 2.56). Based on these numbers we estimate that a five year old child has most likely not seen less than 10.6 and not more than 84.83 million images (total number of fixations during lifetime either scaled down by a factor of 24 or 3).}
\vspace{.2cm}
\footnotesize{\begin{tabular}{ l r r r r r r r r}
\toprule
 Age group & \textit{M} age & Wake time & Fixation duration  & Fixations per second & Min & Lower & Upper & Max \\ 
\midrule
 4--6 year-olds & 5.13 & 99.41 & 390.00 & 2.56 & 4.24 & 10.60 & 84.83 & 254.49\\  
 7--9 year-olds & 8.11 & 154.16 & 380.00 & 2.63 & 6.76 & 16.90 & 135.15 & 405.44\\
 10--12 year-olds & 11.00 & 211.54 & 370.00 & 2.70 & 9.52 & 23.80 & 190.39 & 571.16\\
 13--15 year-olds & 14.22 & 272.42 & 350.00 & 2.86 & 12.99 & 32.43 & 259.43 & 779.12\\
 Adults & 28.33 & 591.71 & 292.42 & 3.42 & 33.73 & 84.32 & 674.55 & 2023.65\\
\bottomrule
\end{tabular}
}
\label{tab:back_envelope}

\end{table}

\begin{table}[!ht]
\centering
\caption{Model details regarding all employed models. \emph{Dataset size} refers to the number of images in the training set and is plotted on the x-axis in Subfigure~\ref{subfig:datasetsize}. \emph{Sample size} is equal to the number of total encountered images during training (dataset size $\times$ epochs) and is plotted on the x-axis in Subfigure~\ref{subfig:samplesize}. *Note that the SWSL model was trained one epoch on 940M images and then 30 epochs on standard ImageNet (1.28M images)---thus sample size equals $940M + 30 \times 1.28M$. \emph{Parameters} refer to the total number of parameters optimised during training and is represented by the area of the circles throughout Figure~\ref{fig:dataset_samplesize}.}

\small{\begin{tabular}{lrrrr}
\toprule
Model   & \multicolumn{1}{l}{Dataset size} & \multicolumn{1}{l}{Epochs} & \multicolumn{1}{l}{Sample size} & \multicolumn{1}{l}{Parameters} \\
\midrule
VGG-19  & 1.28M                            & 74                         & 94.72M                          & 144.00M                        \\
ResNeXt & 1.28M                            & 90                         & 115.20M                         & 44.00M                         \\
BiT-M   & 14.00M                           & 30                         & 420.00M                         & 928.00M                        \\
SWSL    & 940.00M                          & 1+30*                      & 978.40M                         & 829.00M                        \\
SWAG    & 3.60B                            & 2                          & 7.2B                            & 644.80M\\                      
\bottomrule
\end{tabular}}
\label{tab:model_details}
\end{table}

\clearpage

\subsection*{A.5 Shape-bias}
\label{app:shape_bias}

\begin{table}[!ht]
\centering
\caption{Exact fractions  of  texture vs.\ shape  decisions  of  different  age  groups and DNNs in percent.}
\small{\begin{tabular}{ l r r }
\toprule
 Observer & Shape bias & Texture bias \\ 
\midrule
 4--6 year-olds & 87.55 & 12.45 \\  
 7--9 year-olds & 90.86 & 9.14 \\
 10--12 year-olds & 93.52 & 6.48 \\
 13--15 year-olds & 93.18 & 6.82 \\
 Adults & 96.72 & 3.28 \\
 \midrule
 VGG-19 ($>$1M) & 7.96 & 92.04 \\
 ResNeXt ($>$1M) & 25.52 & 74.48 \\
 BiT-M ($>$10M) & 57.24 & 42.76 \\  
 SWSL ($>$100M)& 55.88 & 44.12 \\ 
 SWAG ($>$1,000M) & 45.00 & 65.00 \\ 
\bottomrule
\end{tabular}}
\label{tab:shape_bias}
\end{table}

\begin{figure}[!ht] 
    \begin{center}
    \includegraphics[scale=.32]{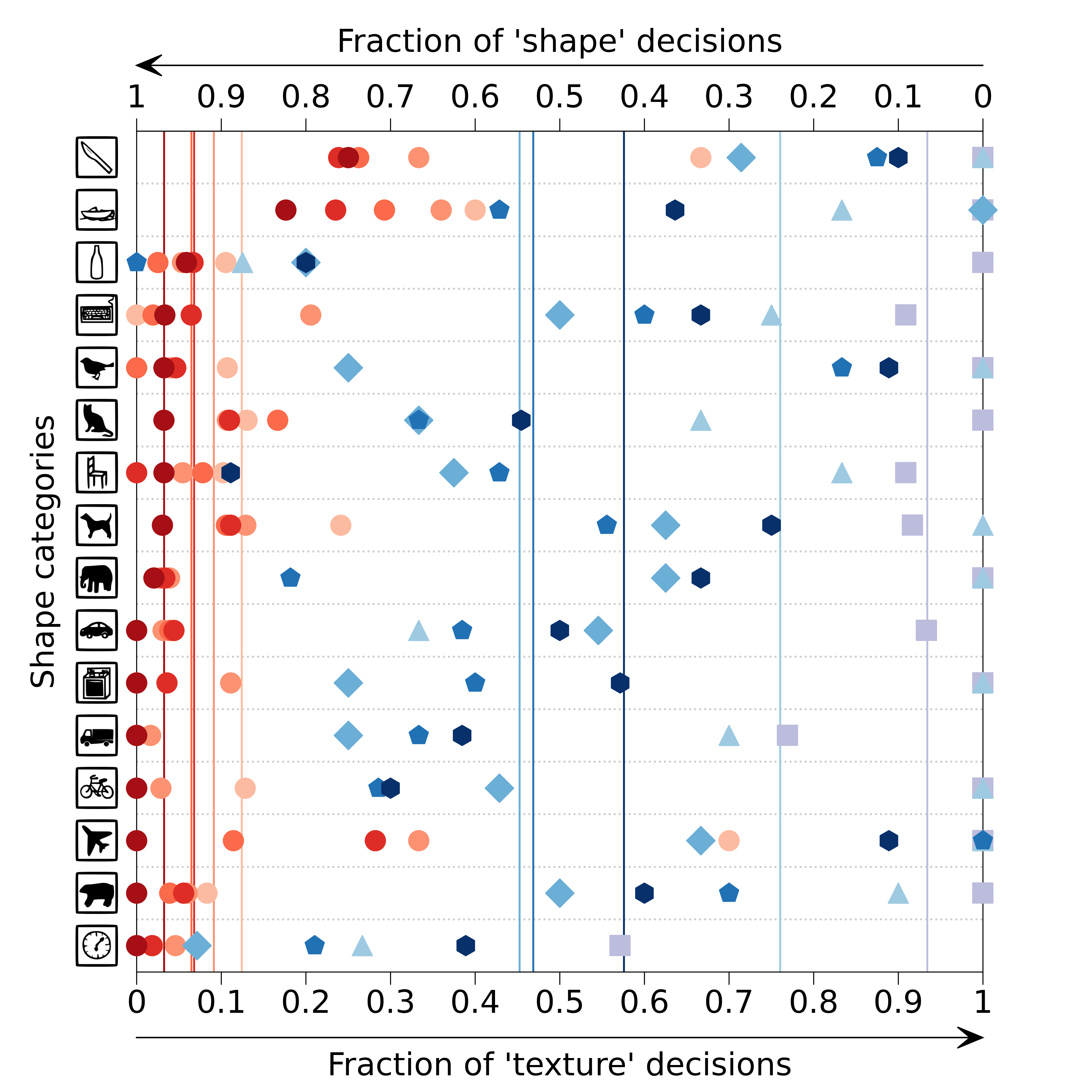}
    \end{center}
    \caption{Category wise proportions of texture vs.\ shape decisions for different age groups (\textcolor{4-6}{4--6}, \textcolor{7-9}{7--9}, \textcolor{10-12}{10--12}, \textcolor{13-15}{13--15}, and \textcolor{adults}{Adults}) and DNNs (\textcolor{vgg19}{VGG-19}, \textcolor{resnext}{ResNext}, \textcolor{bitm}{BiT-M}, \textcolor{swsl}{SWSL}, and \textcolor{swag}{SWAG}). Only responses that corresponded to either the correct texture or correct shape category are taken into account. Vertical lines indicate averages over all categories.}
    \label{fig:shape_bias}
\end{figure}
\clearpage

\subsection*{A.6 Error consistency}
\label{app:errorK}

\begin{figure}[!ht]
    \begin{center}
    
    \includegraphics[scale = .43]{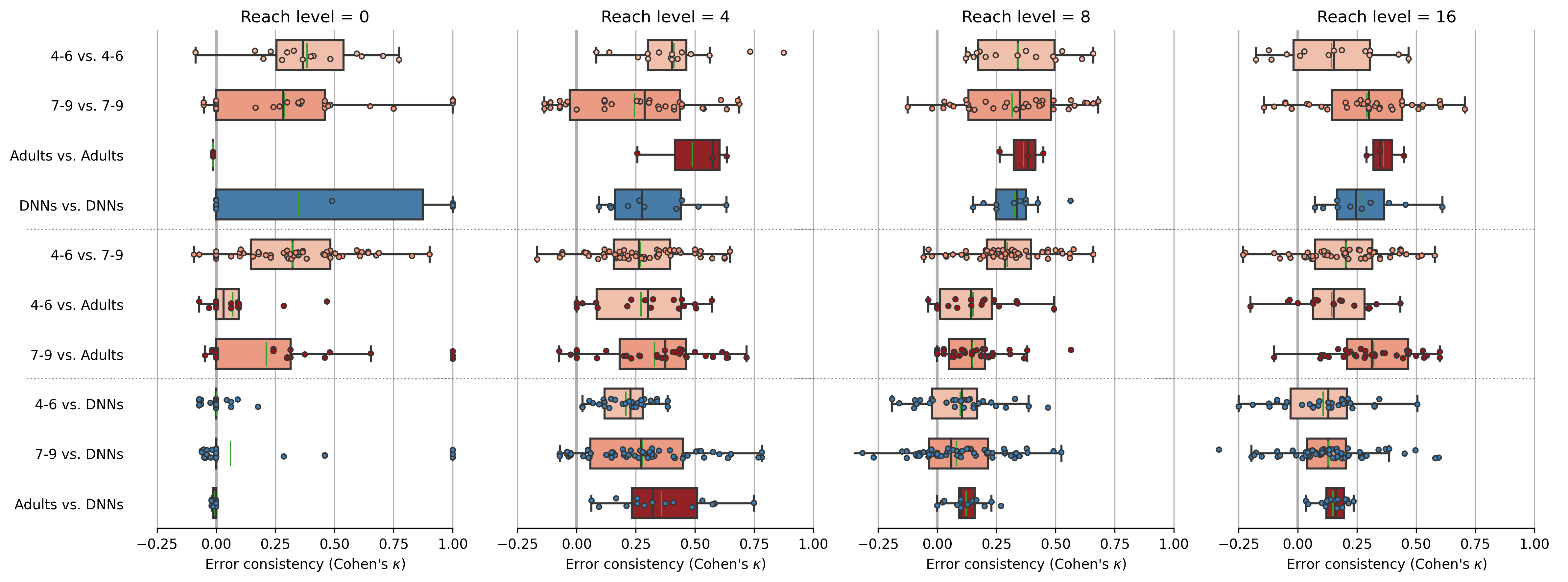}
    \end{center}
    \caption{Error consistency as measured by Cohen's kappa ($\kappa$) for different distortion levels split by different within- and between-group comparisons (\textcolor{4-6}{4--6}, \textcolor{7-9}{7--9}, \textcolor{adults}{adults}, and \textcolor{midblue_errorK}{DNNs}) for the eidolon experiment. $\kappa = 0$ indicates chance level consistency (i.e., both systems are using independently different strategies), $\kappa > 0$ means consistency above chance level (i.e., both systems are using similar strategies) and $\kappa < 0$ means inconsistency beyond chance level (i.e., both systems use inverse strategies). Grid line at $\kappa = 0$ is highlighted. Plots are horizontally divided in three subsection*s: Upper subsection* (within-group comparisons), middle subsection* (between-group comparisons humans only), and lower subsection* (between-group comparison humans and DNNs). Colored dots represent error consistency between two single observers  (one from each group featured in the comparison). Box plots represent the distribution of error consistencies from observers of the two given groups. Boxes indicate the interquartile range ($IQR$) from the first ($Q1$) to the third quartile ($Q3$). While vertical black markers indicate distribution medians, vertical green markers indicate distribution means. Whiskers represent the range from $Q1-IQR$ to $Q3+IQR$. Compared groups are indicated by the color of dots and boxes, respectively.}
    \label{fig:errorK_eidolon}
\end{figure}

\clearpage

\printbibliography

\end{document}